\documentclass[lettersize,journal,twoside]{IEEEtran}
\usepackage{fancyhdr}
\usepackage{amsmath,amsfonts}
\usepackage{algpseudocode}
\usepackage{algorithm}
\usepackage{array}
\usepackage{booktabs}
\usepackage{textcomp}
\usepackage{stfloats}
\usepackage{url}
\usepackage{verbatim}
\usepackage{graphicx}
\usepackage{cite}
\usepackage{caption}
\usepackage{subcaption}
\usepackage{multirow}

\usepackage{lipsum}
\usepackage{pgfplots}
\pgfplotsset{compat=newest}
\usepackage{subcaption}
\usepackage{xcolor}
\colorlet{myyellow}{yellow!50!orange}
\colorlet{mygreen}{green!60!black}

\begin{document}


\title{Marking the Pace: A Blockchain-Enhanced Privacy-Traceable Strategy for Federated Recommender Systems}

\author{Zhen Cai,~\IEEEmembership{Student Member,~IEEE}, Tao Tang,~\IEEEmembership{Student Member,~IEEE}, Shuo Yu,~\IEEEmembership{Member,~IEEE}, Yunpeng Xiao, and Feng Xia,~\IEEEmembership{Senior Member,~IEEE}

\thanks{Manuscript received November 15, 2022; revised June 19, 2023; accepted October 16, 2023. This paper is partially supported by XPCC Science and Technology Planning Project No. 2021AA006 and the ``High-level Talent Team" Project of Dalian Science and Technology Talent Innovation Support Policy Program under Project No. 2022RG11. \textit{(Corresponding author: Shuo Yu)}.}
\thanks{Zhen Cai and Tao Tang are with the Institute of Innovation, Science, and Sustainability, Federation University Australia, Ballarat, VIC 3353, Australia (email: zhen.cai@ieee.org, tao.tang@ieee.org).}
\thanks{Shuo Yu is with the School of Computer Science and Technology, Dalian University of Technology, Dalian 116024, China, and with the Key Laboratory of Social Computing and Cognitive Intelligence (Dalian University of Technology), Ministry of Education, Dalian 116024, China (e-mail: shuo.yu@ieee.org).}
\thanks{Yunpeng Xiao is with the Institute of Electronic Information and Network Engineering, Chongqing University of Posts and Telecommunications, Chongqing 400065, China (e-mail: xiaoyp@cqupt.edu.cn).}
\thanks{Feng Xia is with the School of Computing Technologies, RMIT University, Melbourne, VIC 3000, Australia (e-mail: f.xia@ieee.org).}}



\maketitle

\newcommand{\shortauthors}{Cai et al.}
\newcommand{\shorttitle}{Marking the Pace: A Blockchain-Enhanced Privacy-Traceable Strategy for Federated Recommender Systems}

\begin{abstract}






Federated recommender systems have been crucially enhanced through data sharing and continuous model updates, attributed to the pervasive connectivity and distributed computing capabilities of Internet of Things (IoT) devices. Given the sensitivity of IoT data, transparent data processing in data sharing and model updates is paramount. However, existing methods fall short in tracing the flow of shared data and the evolution of model updates. Consequently, data sharing is vulnerable to exploitation by malicious entities, raising significant data privacy concerns, while excluding data sharing will result in sub-optimal recommendations. To mitigate these concerns, we present LIBERATE, a privacy-traceable federated recommender system. We design a blockchain-based traceability mechanism, ensuring data privacy during data sharing and model updates. We further enhance privacy protection by incorporating local differential privacy in user-server communication. Extensive evaluations with the real-world dataset corroborate LIBERATE's capabilities in ensuring data privacy during data sharing and model update while maintaining efficiency and performance. Results underscore blockchain-based traceability mechanism as a promising solution for privacy-preserving in federated recommender systems.




\end{abstract}

\begin{IEEEkeywords}
Federated recommender systems, privacy-preserving, traceability, blockchain.
\end{IEEEkeywords}
%
\section{Introduction}

The rapid expansion and penetration of Internet of Things (IoT) devices have supercharged recommender systems, facilitating enhanced data sharing and continuous model updates through the harnessing of pervasive connectivity and distributed computing capabilities~\cite{zhou2020deep, chen2023privacy}. Current studies in diverse domains, including health informatics, smart homes, and smart automobiles, have explored the potential of integrating recommender systems into IoT environments~\cite{zhou2020cnn, wu2022eagcn, zhou2023hierarchical, zhou2019multi}. Leveraging recommender systems within federated learning has surfaced as a novel approach that enables multiple devices to collaboratively train a model while preserving privacy \cite{zhou2023decentralized}. Within the context of federated recommender systems, data sharing and model updates emerge as an instrumental function. The practice of data sharing improves the recommendation model’s performance by mitigating data heterogeneity challenges \cite{zhao2018federated, li2019convergence}. Continuous model updates, utilizing diversified IoT device data, further enhance the model’s resilience \cite{su2022boost, huang2022stochastic}.


Transparent is paramount in federated recommender systems. Especially when considering data collected from IoT devices like smart wearable devices, which often contain highly sensitive information~\cite{elayan2021sustainability}. Due to the recent increased attention towards ethical principles governing data-driven decision-making tools such as the General Data Protection Regulation (GDPR)\footnote{https://gdpr-info.eu/} and Australia's AI Ethics Principles\footnote{https://www.industry.gov.au/publications/australias-artificial-intelligence-ethics-framework/australias-ai-ethics-principles}, ethical principles, especially transparency, have received more attention than the prediction accuracy in recommender systems~\cite{fan2022comprehensive, cai2021reliable}. The lack of transparency may impede user's confidence and acceptance of such system.


However, existing mechanisms fall short of tracing the flow of shared data and the evolution of model updates, raising significant concerns about data privacy. Within data sharing, this gap obfuscates the provenance and diffusion paths, making users' data susceptible to misuse by malicious entities~\cite{bhatt2020explainable, himeur2022blockchain}. Furthermore, federated recommender systems are prone to inference attacks from both server and client side~\cite{nasr2019comprehensive}. Malicious server or clients may rely on the updated models to identify other users training data~\cite{wang2021fast, hu2022membership}. The decentralized nature of federated recommender systems complicates the detection of malicious activities or users, thereby exacerbating privacy-related risks~\cite{li2020learning}.

Existing efforts to address data privacy concerns in federated recommender systems primarily rely on differential privacy~\cite{wang2020achieving, zhao2020local, wu2021fedgnn} and homomorphic encryption~\cite{yao2019recommendations, yang2020federated}. Differential privacy achieves privacy-preserving by introducing random noise on updated model parameters~\cite{abadi2016deep, truex2020ldp, zhang2019pefl}. Despite the fact that the computational overhead of implementing differential privacy is relatively small, the introduced noises unavoidably affect model performance, particularly at higher privacy budgets~\cite{wei2020federated, adnan2022federated}. Conversely, homomorphic encryption secures model parameters by encrypting it prior to communication, ensuring that model parameters remain encrypted until decryption~\cite{wood2020homomorphic, fang2021privacy}. Deploying homomorphic encryption techniques for privacy-preserving in federated recommender systems will not compromise model's performance~\cite{li2021survey, kaissis2020secure}. Although this preserves model performance, the significant computational expenses and time required for encryption and decryption make it impractical for scenarios involving frequent data processing~\cite{ma2022privacy}. Additionally, current studies of providing reasonable privacy guarantees mainly focus on the raw data and the model parameters, omitting the non-transparent nature of the whole system. Therefore, lacking traceability, the recommendation process could suffer malicious manipulation~\cite{zhang2020gan, zhang2022pipattack, ma2022shieldfl}, further posing trust related risks~\cite{wu2021triple, himeur2022latest}. Hence, it is significant to design reliable traceability mechanisms to prevent the aforementioned risks.

In this study, we propose LIBERATE, an privacy-traceable federated recommender system. We design a blockchain-based traceability meachanism to address privacy concerns during data sharing and model updates within federated recommender systems. The proposed framework is comprised of two main modules: data storage and processing module, federated matrix factorization training module. Within the data storage and processing module, blockchain-based traceability mechanism is deployed to record the data sharing history among users in the IoT and model updates. This approach leverages the immutable, secure, and fully decentralization nature of blockchain technology, enabling users to trace their data usage and model updates. We further employ local differential privacy in the training module to perturb the uploaded gradient data from users, enhancing privacy protection during communication without the need for third-party service providers. Notably, our approach represents the first effort to ensure privacy in data sharing and model updates utilizing blockchain-based traceability within the context of federated recommender systems in IoT, thereby addressing concerns about data privacy. The main contributions of this paper are listed below:

\begin{itemize}

\item \textbf{Privacy-traceable Federated Recommender System:} We propose a privacy-traceable federated recommender system, namely LIBERATE, which can provide reliable traceability of users' privacy data without compromising the model efficiency and performance.
\item \textbf{Blockchain-enhanced Traceability Strategy:} We design a novel blockchain-enhanced traceability Strategy, leveraging the unique immutable, secure, and fully decentralization nature features of blockchain for record data sharing and continuous model updates during the training process of federated recommender systems.


\item \textbf{Exceptional Performance:} We conduct comprehensive experiments on real-world dataset. Experimental results demonstrate LIBERATE's exceptional privacy-preserving capabilities in data sharing and model updates without compromising model performance or efficiency. 


\end{itemize}





The rest of this paper is structured as follows. Section~\ref{sec2} discusses several related studies on blockchain, traceability, and federated recommender systems. Section~\ref{sec3} introduces preliminaries and background knowledge of our study. Section~\ref{sec4} explains the overall system design. The experimental results of model performance, communication efficiency as well as privacy analysis are discussed in Section~\ref{sec5}. We then summarise our study in Section~\ref{sec6}. 

\section{Related Work}\label{sec2}
This section provides an overview of existing works highly related to transparency in recommender systems and matrix factorization-based recommender systems. 
\subsection{Transparency in Recommender Systems}
Transparency is a paramount principle emphasized by legal regulations governing data-driven tools, such as the GPDR and Australia's AI Ethics Principles. These principles and regulations stipulated that mechanisms of transparency and responsible disclosure should be in place, empowering individuals to understand when and how they engage with data-driven tools. 
In recommender systems, a common way of achieving transparency is to provide explainable recommendations, which aim at enhancing the transparency of the recommendation process and providing more persuasive results for the participants~\cite{bhatt2020explainable}. Wu et al.~\cite{wu2019context} applied an attention mechanism to spotlight important words, thereby offering explainable suggestions to end-users. Similarly, Chen et al.~\cite{chen2019personalized} proposed a model to generate visually explainable recommendations for users within the fashion industry. Ren et al.~\cite{ren2017social} enhanced recommendation explainability via a social collaborative viewpoint regression model.

However, while these explainable mechanisms ensured transparency in recommendation outcomes, the non-transparent nature of those mechanisms presented potential concerns. This opacity might cause malicious recommenders to manipulate explanation results when a recommender system lacks a higher level of transparency. Hence, a novel solution, traceability mechanism, has gained lots of attention in recent years. Previous efforts on achieving traceability of data-driven systems commonly implemented into traceable supply chains by using the tracking information, such as label information~\cite{sunny2020supply} and position information of GPS~\cite{kandel2011gps}. Besides, inspired by the advanced traceability of blockchain schemes, traceable data-driven systems have also been deployed for pharmacy distribution~\cite{9314090}, financial loan management~\cite{wang2019loc}, and agriculture production trace~\cite{salah2019blockchain}.

Current studies on blockchain-based recommender systems mainly focus on enhancing personal information security~\cite{lisi2019smart} and privacy protection~\cite{frey2016collaborative} by implementing blockchain schemes on the computation of recommendation results based on data from E-commerce, IoT networks, E-learning and social networks~\cite{himeur2022blockchain}. 
In the decentralized recommender systems, several efforts have been made to achieve traceability by using blockchains. Bosri et al. \cite{bosri2020integrating} proposed a blockchain-integrated recommender system that recorded the details of data usage of the platform. Users could access the blockchain to trace the transaction history of data usage and even receive incentives. Furthermore, Li et al.~\cite{li2019blockchain} introduced a consensus mechanism to incentivize users while verifying computed gradient data. They used blockchain to log the smart contract for incentive distribution and the uploaded gradient data, thereby enhancing server transparency and trustworthiness. However, implementing smart contracts and validation within the consensus mechanism was time-intensive due to computational complexity. Wang et al. \cite{wang2020trusted} employed blockchain to chronicle the training model and recommendation information for permanent storage, ensuring system transparency. They also adopted differential privacy to secure communication. Yet, their framework did not sufficiently guarantee transparency in data sharing amongst interconnected IoT devices.

Therefore, blockchain schemes could serve as a promising solution to facilitate traceability for enhancing the transparency of a recommender system and further alleviating the privacy concerns of the end-users. In our paper, we choose the 256-bit secure hash algorithm for logging the details of data-sharing status, thereby achieving the traceability of our proposed LIBERATE recommender system.

\subsection{Matrix Factorization-based Recommender Systems}

Matrix factorization has been progressively receiving attention in the machine learning domain \cite{cai2010graph}. It was found successful implementations in diverse applications, spanning from recommendation services to environmental monitoring ~\cite{wang2016sparse, koren2009matrix}. The majority of the studies related to matrix factorization had employed differential privacy and cryptographic strategies to safeguard user privacy against potential breaches.


Berlioz et al. \cite{berlioz2015applying} incorporated local differential privacy to obfuscate original user data prior to its submission to the server for training, thereby ensuring privacy protection. Despite the feasibility of local differential privacy in preserving privacy, it might potentially compromise the performance of the model, highlighting a trade-off between privacy and accuracy. On the contrary, Kim et al. \cite{kim2016efficient} protected user privacy through full homomorphic encryption. The proposed framework involved a third-party cryptographic service server to encrypt user data pre-upload. Despite this method not impacting the model's performance, the additional time requirement for the encryption and decryption processes can result in delayed and inefficient recommendations. Furthermore, the assumed security and trustworthiness of the third-party encryption service provider may lead to privacy breaches if it colludes with the recommendation service provider, thereby rendering homomorphic encryption a potential risk.

Decentralized machine learning approaches like federated learning were combined with the matrix factorization model to bolster its privacy-preserving capability \cite{chai2020secure, huang2022geographical}. Chai et al. \cite{chai2020secure} introduced a federated matrix factorization model that merged federated learning with the matrix factorization model, securing user privacy during the training phase. The model further incorporated full homomorphic encryption for additional user privacy protection during user-server communication \cite{zhu2019deep}. However, the efficiency of the proposed study was questionable due to the time consumption induced by the application of encryption techniques.

The matrix factorization model has a high interpretability of naturally occurring data. It has the ability to capture implicit relationships between users and items. The inherent simplicity of the matrix factorization framework not only ensures computational efficiency but also allows for intuitive interpretations of latent factors. Moreover, its parallelizable nature makes it scalable for large datasets. There, we have employed the matrix factorization model as the foundation for our recommender system.

\section{Preliminaries}\label{sec3}

This section demystifies the preliminaries and background knowledge related to our study. Notations and descriptions are presented in Table \ref{tab:notations}.


\subsection{Federated Matrix Factorization}

Federated matrix factorization is a decentralized approach to collaborative filtering \cite{chai2020secure}. The objective is to converge user profile matrix \(U\) and item profile matrix\(V\). Considering \(m\) users and \(n\) items, the prediction of user \(i\)’s rating \(r_{ij}\) on item \(j\) is articulated through the dot product of \(u_i\) and \(v_j\), where \(u_i\) resides in matrix \(U\) and \(v_j\) in matrix \(V\), with \(i\) in set \(\{1, 2, 3, \ldots, m\}\) and \(j\) in set \(\{1, 2, 3, \ldots, n\}\).

The training process initiates by configuring an \(n \times l\) matrix \(V\) and an \(m \times l\) matrix \(U\) at a centralized server and across each client, respectively. The matrix \(V\) is then distributed to each user \(i\), facilitating local training through private dataset \(R_i\). This enables the computation of gradient data $\nabla_{u_{i}} w_{ui}$ and $\nabla_{v_{i}} w_{vi}$ for local matrix \(U\) and the global matrix \(V\), respectively. Each user sends the $\nabla_{v_{i}} w_{vi}$ back to the central server. The server then aggregates the received gradient data and updates the item profile matrix \(V\) accordingly. The updated matrix \(V\) is then used for further iterations.

\begin{table}[h]
\centering
\caption{Summary of Main Notations.}
\label{tab:notations}
\begin{tabular}{cp{6cm}}
\toprule
\textbf{Notation} & \textbf{Explanation} \\
\midrule
$m$ & Number of users \\
$n$ & Number of items \\
$M$ & Total number of ratings \\
$l$ & Dimension of the user/item profile matrix \\
$U$ & $m \times l$ user profile matrix \\
$V$ & $n \times l$ item profile matrix \\
$R$ & $m \times n$ rating matrix \\
$r_{ij}$ & Rating by user $i$ for item $j$ \\
$u_{i}$ & User $i$'s user profile matrix \\
$v_{i}$ & User $i$'s item profile matrix \\
$\lambda$ & Regularization parameter \\
$\gamma$ & Learning rate \\
$\nabla_{u_{i}} w_{ui}$ & Gradient with respect to $u_{i}$ \\
$\nabla_{v_{i}} w_{vi}$ & Gradient with respect to $v_{i}$ \\
$LDP$ & Local differential privacy noise \\
\bottomrule
\end{tabular}
\end{table}
\subsection{Local Differential Privacy}

Local differential privacy emerges as a promising solution for privacy-preserving attributed to its inherent computational efficiency and autonomy from reliance on third-party intermediaries. By definition, the mechanism $P$ that satisfies differential privacy is called $\epsilon$-differentially private when: 
\begin{equation}
\operatorname{Pr}[P(D) \in S] \leq e^{\epsilon} \operatorname{Pr}\left[P\left(D^{\prime}\right) \in S\right],
\end{equation}
where ${D}$ and ${D^{\prime}}$ denote two identical adjacency datasets with one record in difference. ${S}$ denotes all possible outputs. The $\epsilon$ is the privacy budget that controls the level of interference of the noise. In our model, 0 mean Laplace noise is added to the uploaded gradients to avoid information leakage as the server is assumed to be curious. The Laplace differential privacy algorithms we used in this model is defined as: \textit{In a given function f: $D^{n} \rightarrow Y$, where $\epsilon$ is greater than 0 and Y is the set of all outcomes, the equation of the Laplace mechanism is defined as follows:}
\begin{equation}
P(D)=f(D)+\operatorname{Lap}\left(0, \frac{\Delta_{f}}{\epsilon}\right).
\end{equation}

\subsection{Blockchain}


\begin{figure}[h]
\centering
\captionsetup{justification=centering}
\includegraphics[scale = 0.19]{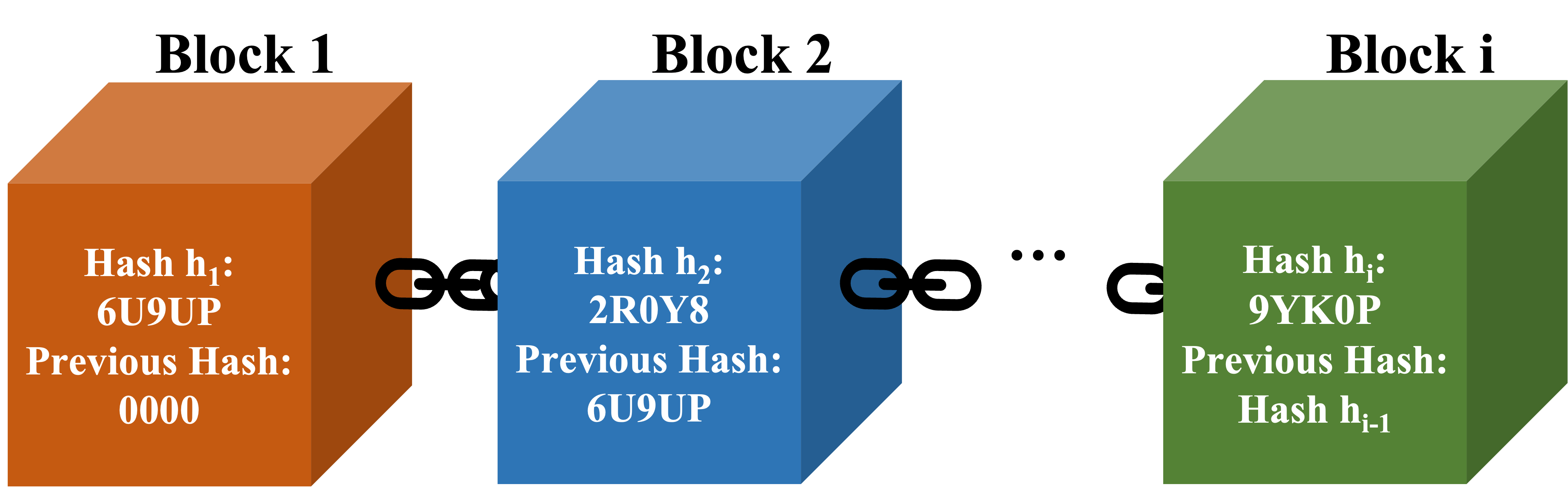}
\caption{An illustrative example of a blockchain: each block's unique hash value serves to authenticate and maintain the integrity of its place within the chain.}
\label{fig:bcg}
\end{figure}

\begin{figure*}[!b]
\centering
\captionsetup{justification=centering}
\includegraphics[scale = 0.5]{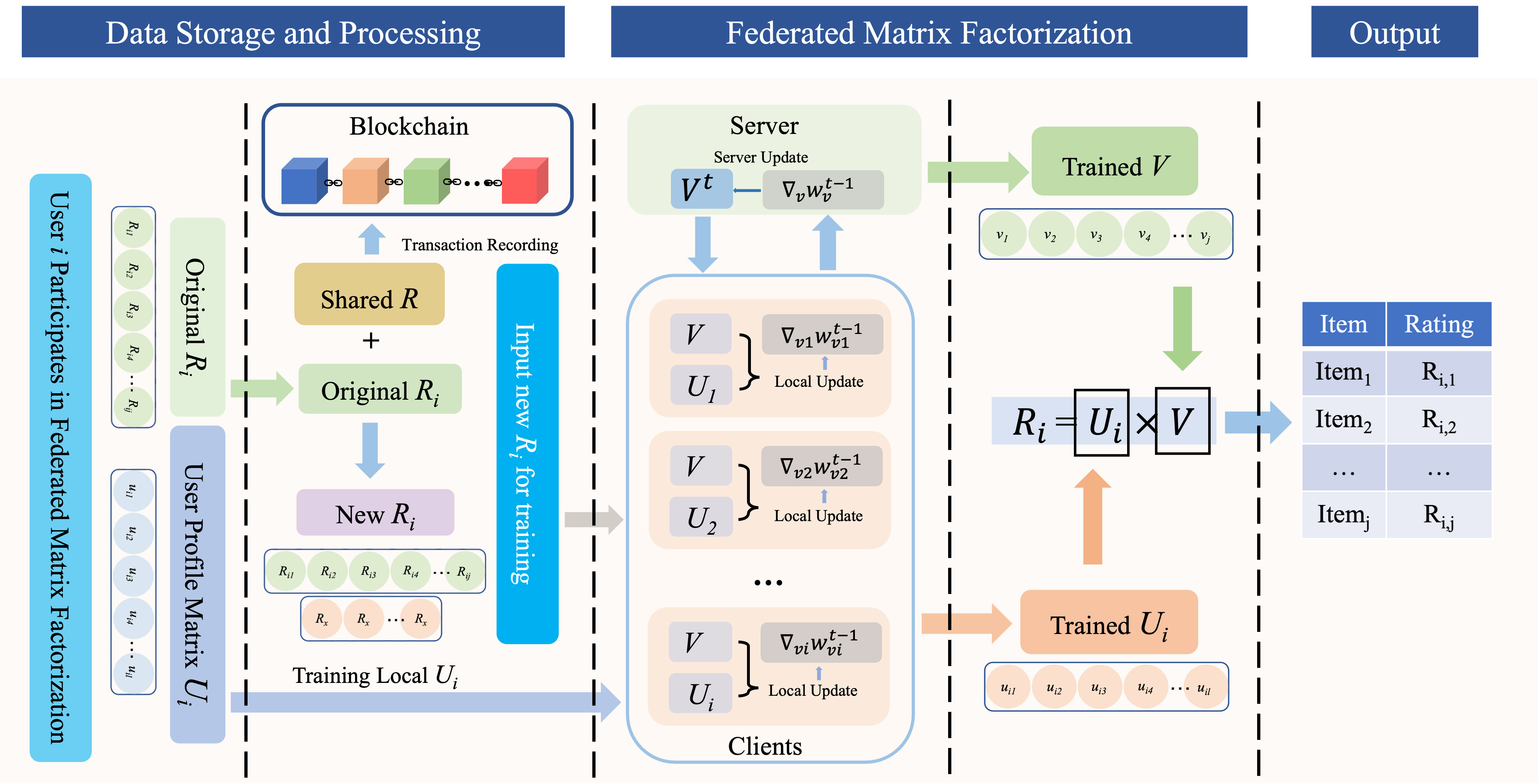}
\caption{An overview of LIBERATE.}
\label{fig:model}
\end{figure*}

Blockchain stands as a potentially in enhancing traceability across various fields like supply chain and healthcare management ~\cite{sunny2020supply, singh2022framework}. As exemplified in Fig. \ref{fig:bcg}, blockchain establishes a sequential and interconnected block structure. Each block encapsulates a recorded hash value pivotal for ascertaining the integrity of the encompassed data within the chain. Melding these two critical aspects, the implementation of blockchain technology can significantly amplify the privacy-preserving of data sharing and model updates in federated recommendation systems.

In LIBERATE, we leverage the traceability of blockchain for privacy-preserving. Specifically, the blockchain-based traceability mechanism records every updates in a secure and immutable log, capturing details such as the rating matrix $R$, as well as the user and item profile matrices $U$ and $V$. This ensures the entire history of data sharing and model updates is permanently and transparently documented. Furthermore, this record is made accessible to all participants. Thereby facilitating the privacy-preserving of data sharing and model updates.



\subsection{Problem Definition}

Given a federated recommendation system, the objective is to achieve privacy-preserving in data sharing and model updates. The sharing of the rating matrix $R$, updates to the user profile matrix $U$ and item profile matrix $V$ are meticulously recorded on blockchain-based traceability mechanism. Ensuring that each transaction and modification is accurately attributed. Thus alleviating concerns related to data privacy during data sharing and model updates.

\section{The Design of LEBERATE}\label{sec4}

This section discusses the overall system architecture and platform design of LIBERATE, which comprises two main modules: the data storage and processing, and the federated matrix factorization. LIBERATE incorporates blockchain-based traceability mechanism to ensure privacy-preserving in data sharing within the data storage and processing module. Additionally, this mechanism is employed in the training module to record the updated user profile matrix $U$ and item profile matrix $V$. The training model follows a federated matrix factorization structure. To ensure communication privacy and time efficiency, we implemented local differential privacy techniques during the upload of gradient data, providing an additional layer of protection against information leakage. The result calculation module leverages the converged user profile matrix $U$ and item profile matrix $V$ from the training module to generate final recommendation results for users. Fig. \ref{fig:model} provides an overview of the system structure.

\begin{figure*}[!t]
\centering
\captionsetup{justification=centering}
\includegraphics[scale = 0.55]{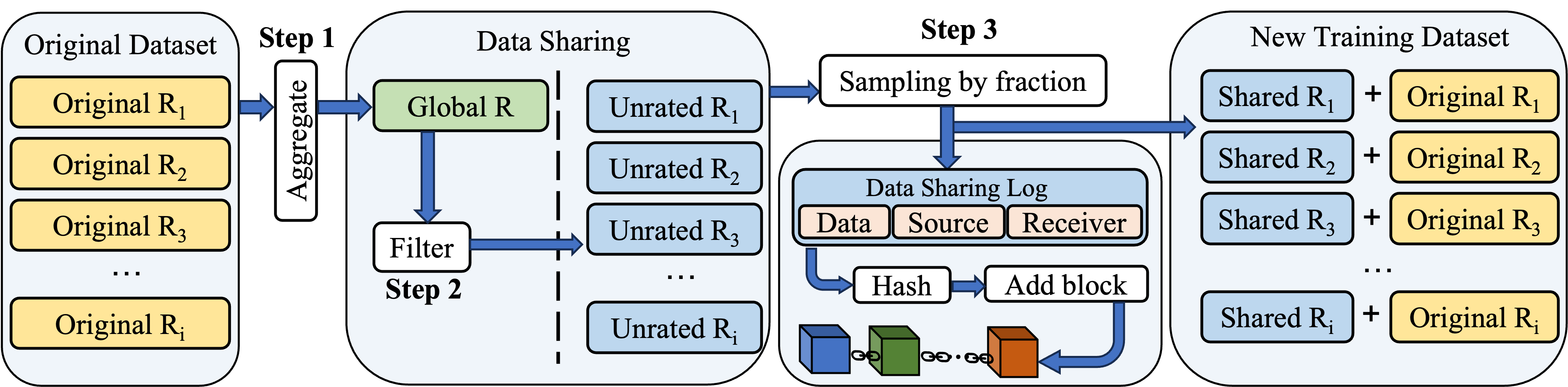}
\caption{The illustration of data storage and processing module.}
\label{fig:ds}
\end{figure*}

\subsection{Data Storage and Processing}


In the data storage and processing module, we utilize blockchain-based traceability mechanism to record the data sharing history between interconnected devices. By harnessing the inherent characteristics of invariance, security, and full distribution offered by blockchain, our approach ensures the privacy-preserving of data sharing activities.

Whenever a user extends their local dataset by receiving randomly sampled datasets from other users, a new block is created in the blockchain to log the pertinent data sharing details. These data sharing logs encompass crucial information, including the data source, receiver, timestamp, shared dataset, and the hash of the block. Furthermore, the hash of the previous block is included to validate the integrity and continuity of the blockchain. To provide clarity, Fig. \ref{fig:ds} presents an illustrative example depicting the steps involved in the data sharing processing. The following steps outline the data sharing procedure:
 
\textbf{Step 1.} Users upload their local rating data to a global database $R$.

\textbf{Step 2.} The unrated item of user $i$ are filtered out as unrated $R_i$.

\textbf{Step 3.} User $i$ will then receive a fraction of rating data from the unrated $R_i$, combine with the original $R_i$.

\textbf{Step 4.} The user then starts training with the new training dataset.

Since blockchain is tamper-proof and the transaction history in each block cannot be modified once it is created, the flow of data provided and received by each user can be recorded to facilitate future tracking of data flow and finding the user responsible for the data. Additionally, the user profile matrix $U$ and item profile matrix $V$ are recorded, enabling the privacy-preserving of model updates. Algorithm \ref{alg:blockchain} details the blockchain-based traceability mechanism utilized in our proposed LIBERATE system.

\begin{algorithm}[]
\caption{Blockchain-based Traceability Mechanism}\label{alg:blockchain}
\begin{algorithmic}[1]
\Require{$\text{Data Sharing Logs}$, $U$, $V$}
\Ensure{Blockchain with recorded data}

\State Initialize an empty blockchain $\text{BC}$

\For{$\text{each data sharing log}$}
    \State $\text{hash} \gets \text{H}(\text{data sharing log})$
    \State $\text{BC}.\text{addBlock}(\text{hash})$
\EndFor

\State $\text{hashUser} \gets \text{H}(\textit{U})$
\State $\text{BC}.\text{addBlock}(\text{hashUser})$

\State $\text{hashItem} \gets \text{H}(\textit{V})$
\State $\text{BC}.\text{addBlock}(\text{hashItem})$

\State \Return $\text{Blockchain}$
\end{algorithmic}
\end{algorithm}

To maintain the integrity and security of the recorded data within the blockchain, we employ the Secure Hash Algorithm 256-bit (SHA256) hash function \cite{yaga2019blockchain}. The SHA-256 is a cryptographic hash function that produces a fixed-size output of 256 bits, which is approximately 64 characters in hexadecimal representation, from an input of arbitrary length. Mathematically, the SHA-256 hash function can be represented as:
\begin{equation}
H: \{0,1\}^* \rightarrow \{0,1\}^{256},
\end{equation}
where \(H\) is the hash function. The domain of \(H\) represents the set of all binary strings of arbitrary length, which, in the context of our application, includes representations of data-sharing logs, user profiles \(U\), and item profiles \(V\). The co-domain is the set of 256-bit binary strings. One of the primary properties of SHA-256 is that it is collision-resistant, which means it is computationally infeasible to find two different inputs that produce the same output. This property is mathematically expressed as:

\begin{equation}
\forall x_1, x_2 \in \{0,1\}^* : x_1 \neq x_2 \Rightarrow H(x_1) \neq H(x_2)
\end{equation}

Additionally, SHA-256 is deterministic, implying that the same input will always produce the same output. This hash function also exhibits the avalanche effect, where a small change in the input will produce a drastically different output. This property ensures that even minor variations in the input data (e.g., sharing logs or model updates) will yield entirely distinct hash values, facilitating effective traceability in blockchain mechanisms.

\subsection{Federated Matrix Factorization}

\begin{figure*}[t]
\centering
\captionsetup{justification=centering}
\includegraphics[scale = 0.45]{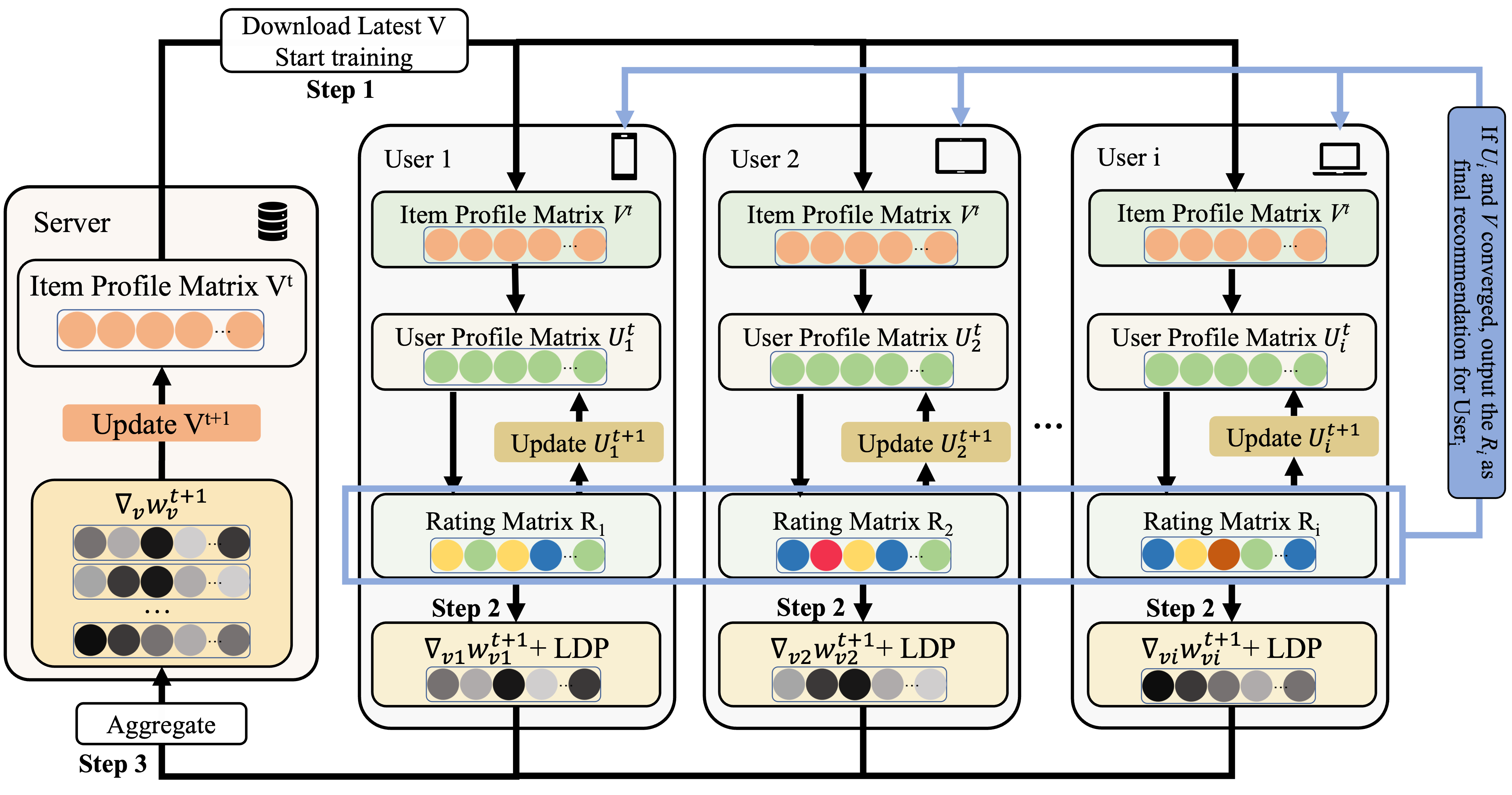}
\caption{Federated matrix factorisation of LIBERATE}
\label{fig:fmfp}
\end{figure*}

After data storage and processing, users can then start training the federated matrix factorization recommendation model. An illustrative example of federated matrix factorization in LIBERATE is displayed in Fig. \ref{fig:fmfp}. In LIBERATE, the server preserves the latest item profile matrix $V$ for the user to download, so that it performs local training to update the user profile matrix $U$. The calculated gradient data from user $u_i$ are then sent back to the server to perform the update on the item profile matrix $V$ and prepared for the next round of training. The objective function is to converge \textit{U} and $V$. The algorithm of the federated matrix factorization is displayed in Algorithm \ref{alg:fedMF}.

\begin{algorithm}[hbt!]
\caption{Federated Matrix Factorization in LIBERATE}\label{alg:fedMF}
\begin{algorithmic}[1] 
\Require{Server preserves $V$; each user $i$ preserves a local $u_i$.}
\Ensure {Optimized $U$ and $V$ to generate recommendations for each user $i$.}
\State Initialize $u_i$ for each user $i$ and $V$ on the server.
\For {$t$ in $iterations$}
\State \textit{\textbf{On each User $i$:}}
\State Download $V$ from the server.
\State Compute local $\nabla_{v_{i}} w^{t-1}_{vi}$ according to Equation (7).
\State Update $u_{i}$ according to Equation (4).
\State Clip the $\nabla_{v_{i}} w^{t-1}_{vi}$ with $LDP$ via Equation (2).
\State Send new $\nabla_{v_{i}} w^{t-1}_{vi}$ to the server.
\State \textit{\textbf{On the Server:}}
\State Receive ${\nabla_{v_{i}} w^{t-1}_{vi}}$ from all users.
\State Aggregate received ${\nabla_{v_{i}} w^{t-1}_{vi}}$.
\State Update $V$ according to aggregated ${\nabla_{v_{i}} w^{t-1}_{vi}}$.
\EndFor

\end{algorithmic}
\end{algorithm}

As the purpose is to optimise the trainable variables \textit{U} and \textit{V}, the computation process can be defined as solving the following regularised least squares minimisation equation:

\begin{equation} 
\min _{U, V} \frac{1}{M}\left(r_{ij}-\left\langle u_{i}, v_{j}\right\rangle\right)^{2}+\lambda(\|U\|_{2}^{2}+\|V\|_{2}^{2}),\end{equation} where M is the total number of ratings, ${r_{i,j}}$ is in the rating matrix $R$ represents the user \textit{i}'s degree of preference for the corresponding item \textit{j}. The ${u_{i}}$ and ${v_{i}}$ represent user \textit{i}'s user and item profile matrix from $U$ and $V$. To predict the ${r_{i,j}}$, we take the ${u_{i}}$ and $j^{th}$ value from ${v_{i}}$. The dot product will be the ${r_{i,j}}$. ${\lambda}$ is a positive value for punisher rescaling purposes. $\lambda(\|U\|_{2}^{2}+\|V\|_{2}^{2})$ denotes L2 regularization element. $U$ and $V$ are updated in each iteration based on the equation defined as follows:
\begin{equation} 
u_{i}^{t}=u_{i}^{t-1}-\gamma \nabla_{u_{i}} w^{t-1}_{ui},
\end{equation}
and 
\begin{equation} 
v_{i}^{t}=v_{i}^{t-1}-\gamma \nabla_{v_{i}} w^{t-1}_{vi},
\end{equation}
where in the ${\nabla_{u_{i}} w^{t-1}_{ui}}$ and ${\nabla_{v_{j}} w^{t-1}_{vj}}$ is defined as: 
\begin{equation}
\nabla_{u_{i}} w^{t-1}_{ui}=-2 \sum_{j:(i, j)} v_{j}\left(r_{ij}-\left\langle u_{i}, v_{j}\right\rangle\right)+2 \lambda u_{i},
\end{equation}
and
\begin{equation}
\nabla_{v_{i}} w^{t-1}_{vi}=-2 \sum_{i:(i, j)} u_{i}\left(r_{ij}-\left\langle u_{i}, v_{j}\right\rangle\right)+2 \lambda v_{i},
\end{equation}

where $\nabla_{v_{i}} w^{t-1}_{vi}$ is represented by $Gradient_i$ within charts. Local differential privacy is further employed to avoid information leakage from gradient data. The server will then aggregate the received $Gradient_i$ from all users to build a $Gradient Matrix$. And therefore, the major steps of federated matrix factorization are as follows:

\textbf{Step 1.} The user \textit{i} first downloads the latest item profile matrix \textit{V} from the server to perform local updates on \textit{U} to compute the ${Gradient_{i}}$.

\textbf{Step 2.} The ${Gradient_{i}}$ then clip with 0 mean Laplace distributed noise to avoid any information leakage from the original ${Gradient_{i}}$.

\textbf{Step 3.} Then, the disturbed ${Gradient_{i}}$ are send to the server to construct a $Gradient Matrix$. This matrix will then be used to update the item profile matrix $V$. And the new $V$ is ready for the next iteration. 

Training of the federated matrix factorization model will cease either upon achieving convergence or when it reaches a predefined level of satisfactoriness, indicative of optimal performance.

After converging the \textit{U} and \textit{V}, the model is ready to predict the recommendation result ${r_{i,j}}$ of an item \textit{j} for user \textit{i} based on the ${\left\langle u_{i}, v_{j}\right\rangle}$ for ${u_{i}}$ in \textit{U} and ${v_{j}}$ in \textit{V} according to the following equation:

\begin{eqnarray}
\left[\begin{array}{c}
r_{11}, r_{12},  \ldots, r_{1 j} \\
r_{21}, r_{22},  \ldots, r_{2 j} \\
r_{31}, r_{32},  \ldots, r_{3 j} \\
\ldots \\
r_{i 1}, r_{i 2},  \ldots, r_{i j}
\end{array}\right] & = & \left[\begin{array}{c}
u_{1} \\
u_{2} \\
u_{3} \\
\ldots \\
u_{i}
\end{array}\right]\times\left[v_{1}, v_{2}, \ldots, v_{j}\right].
\end{eqnarray}

The prediction result is the expected rating of an item by the user. The estimates are then ranked from highest to lowest, providing users with items that are likely to be highly rated for them. After the user has calculated the local recommendation results, \textit{U} and \textit{V} are then stored in the blockchain for tracking and tracing purposes, ensuring a transparent recommendation results computation.

\section{Experiments}\label{sec5}

\subsection{Dataset and Parameter Setting}

In our study, we utilize the MovieLens 1M dataset, comprising ratings from 6,040 users on 3,900 movies \cite{harper2015movielens}. These ratings, ranging between 0 and 5, provide insights into user preferences. This dataset includes an abundance of relational data between users and items, making it an ideal choice for evaluating models focused on relational prediction tasks. To create varying dataset sizes for experimentation, we specifically curated subsets using the top-rated movies and the most active users. Three distinct datasets were formulated by selecting the top 10, 15, and 20 users coupled with the 40, 70, and 90 most-rated movies, respectively. For our experiments, we adopted a split where 80\% of the data was used for training and the remaining 20\% served as the test set.

Due to system capacity constraints, we limited the training to 80 iterations. We set the model's learning rate and the regularization parameter, ${\lambda}$, at ${1e-3}$ and ${1e-4}$, respectively. The dimension of the latent space for the user and item profiles matrix was chosen as 100. Considering the limited availability of training data, we present the differential privacy parameter, $\epsilon$, at 10 and examined the model's performance at varied $\epsilon$ values and in scenarios without differential privacy. The default fraction for data sharing was set at 30 $\%$. To test the model's performance, we conducted 30 rounds of evaluations using randomly sampled datasets for data-sharing and calculated the results using a 95$\%$ confidence interval. These extensive experiments were conducted on a cloud server equipped with a 2.20GHz 6-core CPU and 16GB of RAM. The model is built with Python version 3.9 with Numpy version 1.23. 

\subsection{Evaluation Metrics}
Our evaluation comprises three experiments. The first experiment conducts a comparative analysis of our model's performance metrics. The performance of our model is evaluated based on its proximity to the true ratings based on Root Mean Squared Error (RMSE) and Normalised Discounted Cumulative Gain (NDCG). Specifically, RMSE and NDCG are compared with those of Secure FedMF \cite{chai2020secure}, traditional centralised MF \cite{koren2009matrix}, and AutoRec \cite{sedhain2015autorec}. This analysis provides insights into how our model's performance fares against the benchmarks. Additionally, we analyse the influence of the shared data proportion on the final RMSE outcomes, as well as examine the impact of differential privacy noise introduced during the training phase on the RMSE by adjusting the privacy budget $\epsilon$. We further explore the RMSE trends across datasets of varying sizes encompassing an increasing number of users and items. The second experiment examines the model's time efficiency during interactions between the blockchain and the user-server across different dataset sizes. We also explore the influence of the hash function's difficulty level on the overall time expenditure, thereby providing insights into the time complexity aspects of our model. The final experiment provides a qualitative analysis of LIBERATE's privacy-preserving through detailed case studies. 

\subsubsection{RMSE}RMSE is one of the evaluation metrics we used to evaluate the performance of our model. The RMSE is calculated through:

\begin{equation} 
RMSE = \sqrt{\frac{\sum_{i=1}^{N}\|y(i)-\hat{y}(i)\|^{2}}{N}},
\end{equation}
where in the equation, $N$ represents the number of data, and $y(i)-\hat{y}(i)$ represents the difference between ratings predicted by the model based on the true rating data.

\subsubsection{NDCG}NDCG is yet another evaluation metric widely used to evaluate the quality of recommendation results in the form of ranked lists. The NDCG score is calculated by: 

\begin{equation}
NDCG = \frac{DCG}{iDCG} ,
\end{equation}
where $DCG$ is the discounted cumulative gain, and the $iDCG$ is the discounted cumulative gain of the ideal order of recommendation results. The $DCG$ score is calculated through:

\begin{equation} 
DCG = \sum_{i=1}^{n} \frac{\text { relevance }_{i}}{\log _{2}(i+1)},
\end{equation}

where the $relevance$ of $i$ is the score of each individual item $i$ in a given predicted recommendation list, and $n$ is the total number of recommended items in this list.

\subsection{Model Performance}

\begin{figure}[h]
\hspace{0.3cm}
\centering
\captionsetup{justification=centering}
\includegraphics[width = \linewidth]{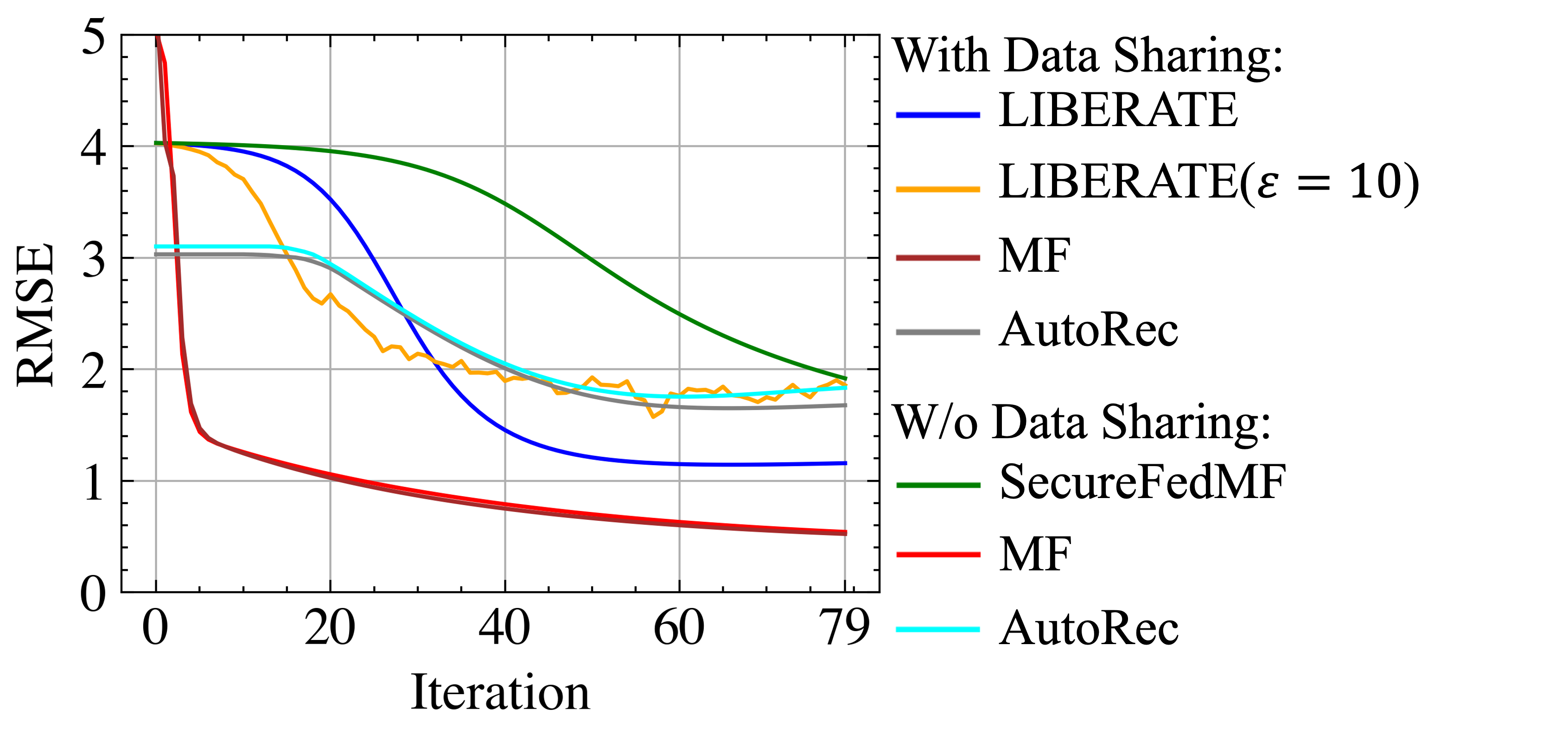}
\caption{Performance benchmarking: RMSE of LIBERATE and baseline models.}
\label{fig:rmse1}
\end{figure}

\begin{table}[h]
\centering
\caption{\label{Tab: rmse} Comparison of RMSE with other models. DP represents differential privacy.}
\begin{tabular}{lcc}
\toprule
\multicolumn{1}{c}{\multirow{2}{*}{Model}} & \multicolumn{2}{c}{RMSE}                 \\ \cline{2-3} 
\multicolumn{1}{c}{}                            & With Data Sharing & Without Data Sharing \\ \midrule
LIBERATE (Without DP)                              & \textbf{1.175 $\pm$ 0.032}             & -                    \\
LIBERATE ($\epsilon$  = 10)                                 & \textbf{1.803 $\pm$ 0.074}            & -                    \\
Secure FedMF                                           & -                 & 1.918                \\
MF                                              & 0.540 $\pm$ 0.001            & 0.543               \\
AutoRec                                         & 1.697 $\pm$ 0.011            & 1.833                \\ \bottomrule

\end{tabular}
\end{table}
The RMSE results of the training process are shown in Fig. \ref{fig:rmse1}. Table \ref{Tab: rmse} presents a comparative evaluation of the RMSE between various models, considering scenarios with and without data sharing. In the context of data sharing, LIBERATE, with and without differential privacy, presents RMSE values of 1.803 $\pm$ 0.074  and 1.175 $\pm$ 0.032, respectively. The MF model presents an impressive performance with an RMSE of 0.540 $\pm$ 0.001, slightly improving without data sharing, demonstrating an RMSE of 0.543. However, as mentioned earlier, the centralized environment of MF poses significant privacy concerns. AutoRec showcases an RMSE of 1.697 $\pm$ 0.011 and 1.833 for the with and without data sharing scenarios, respectively. Due to the similarity in model structure, we only present the result of without data sharing in Secure FedMF. It yields an RMSE of 1.918. In comparison, LIBERATE offers a performant yet privacy solution for recommender systems.

The multilinear plot provides a comparative analysis of the relative performance of LIBERATE and other recommendation models, particularly in terms of RMSE during training. As illustrated in Fig. \ref{fig:rmse1}, LIBERATE notably outperforms other decentralized models in scenarios where differential privacy is not employed. The centralized MF model registers the lowest RMSE, indicating optimal performance. However, the centralized setting might incur potential privacy concerns with user data, casting a shadow on their impressive performance. When differential privacy is introduced, with a designated parameter $\epsilon = 10$, the landscape of performance undergoes a significant change. The RMSE of LIBERATE not only increases but also displays a certain degree of instability throughout the training phase. This instability, or jitter in the RMSE curve, is a direct consequence of the noise introduced by differential privacy. The noise is integrated into the gradient data during user uploads as a privacy-preserving measure, which causes a certain degree of randomness in the learning process. This randomness is manifested as fluctuation in the RMSE curve during the training phase. Despite this introduced variability, LIBERATE's performance remains competitive, especially when compared to AutoRec and Secure FedMF, where user-server communication is left unprotected. This illustrates that LIBERATE's design involves a marginal trade-off of accuracy for significant privacy gains. Therefore, the data suggests that LIBERATE effectively strikes a balance between privacy protection and performance.

It is worth noting that centralized MF demonstrates superior performance in terms of RMSE. This can be attributed to the centralized setting exhibits a lower RMSE due to its direct access to the complete dataset, which ensures learning of both global and local patterns. This centralized approach benefits from optimized, consistent training conditions and avoids the challenges of potential data imbalances seen in decentralized setups. In contrast, the imbalanced data distribution would preventing the model from fully uncovering the patterns in decentralized setting. Imbalances are prevalent in recommender system datasets. Such imbalances can distort the learning process of federated recommender systems, result in overemphasize majority items and neglect minority items. 

\begin{table}[h]
\centering
\caption{\label{Tab: ntcg} Comparison of NDCG with other models. DP represents differential privacy.}
\begin{tabular}{lcc}
\toprule
\multicolumn{1}{c}{\multirow{2}{*}{Model}} & \multicolumn{2}{c}{NDCG}       \\ \cline{2-3} 
\multicolumn{1}{c}{}                       & With Sharing & Without Sharing \\ \midrule
LIBERATE(Without DP)                               & \textbf{0.876 $\pm$ 0.043}        & -               \\
LIBERATE($\epsilon$ = 10)                               & \textbf{0.865 $\pm$ 0.069}        & -               \\
MF                                         & 0.825 $\pm$ 0.012        & 0.799           \\
Secure FedMF                                      & -            & 0.864           \\
AutoRec                                    & 0.819 $\pm$ 0.031        & 0.850           \\ \bottomrule
\end{tabular}
\end{table}

Table \ref{Tab: ntcg} provides a comparative assessment of the NDCG across various models in the scenarios with and without data sharing. As an effective metric for evaluating recommendation systems where the results are generated as a ranked list, NDCG encapsulates the relevance of recommendations to the users.

In the context of data sharing, NDCG results of LIBERATE with differential privacy demonstrate a NDCG scores of 0.865 $\pm$ 0.069, while LIBERATE without differential privacy demonstrate a NDCG scores of 0.876 $\pm$ 0.043. These results signify that LIBERATE maintains a consistently high performance even under the influence of differential privacy noise. MF model exhibits results with an NDCG of 0.825 $\pm$ 0.012, which slightly diminishes to 0.799 when sharing is removed. AutoRec presents NDCG scores of 0.819 $\pm$ 0.031 and 0.850 under the with and without data sharing conditions respectively. Secure FedMF model demonstrates an NDCG score of 0.864, closely competing with the performance of LIBERATE with differential privacy. Altogether LIBERATE showcases superior performance in terms of NDCG across all the compared recommendation models under the data sharing scenario. Notably, the differential privacy version of LIBERATE competes well with other models. The slight drop in NDCG scores with the application of differential privacy in LIBERATE underscores the delicate balance between maintaining privacy and ensuring recommendation performance. The results also highlight the potential for further investigation into the dynamics of data sharing on the recommendation performance, as seen in the case of AutoRec.

Our model includes a data storage and processing module allowing random data sharing among interconnected devices prior to training. The default setting prescribes that the user's local dataset is expanded by 30$\%$ via this sharing process. In an attempt to understand the impact of varying degrees of shared data on the Root Mean Square Error (RMSE) of the model, we undertook tests under data sharing fractions of 10$\%$, 20$\%$, and 30$\%$, with the differential privacy parameter $\epsilon$ uniformly set to 10 across these scenarios.

\begin{figure*}[h]
    \centering
    \begin{minipage}{0.9\linewidth}
        \begin{subfigure}[c]{0.33\textwidth}
            \centering
            \includegraphics[width=\linewidth]{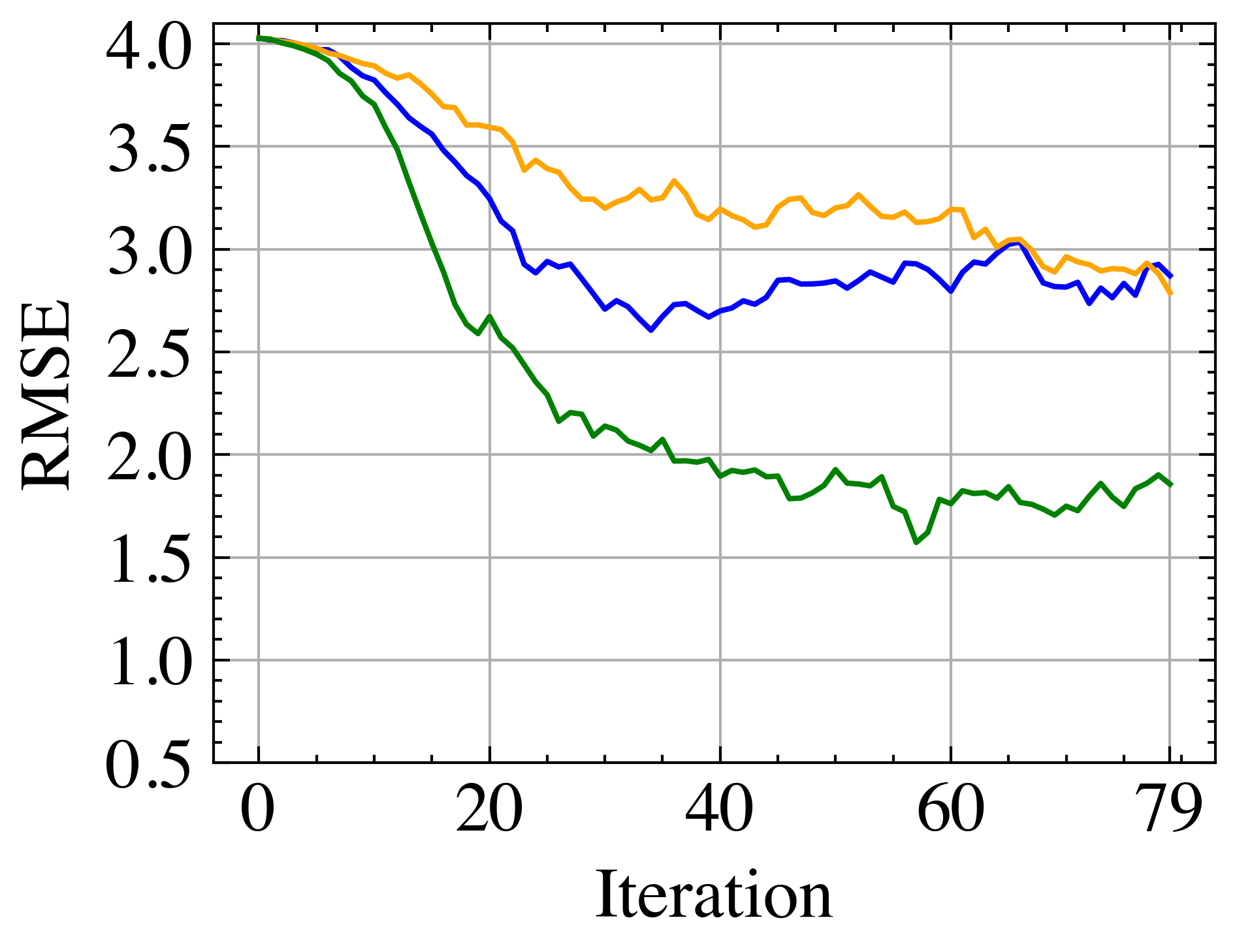}
            \caption{Item = 40, User = 10}
            \label{fig:s40}
        \end{subfigure}
        \hspace{-0.2cm}
        \begin{subfigure}[c]{0.33\textwidth}
            \centering
            \includegraphics[width=\linewidth]{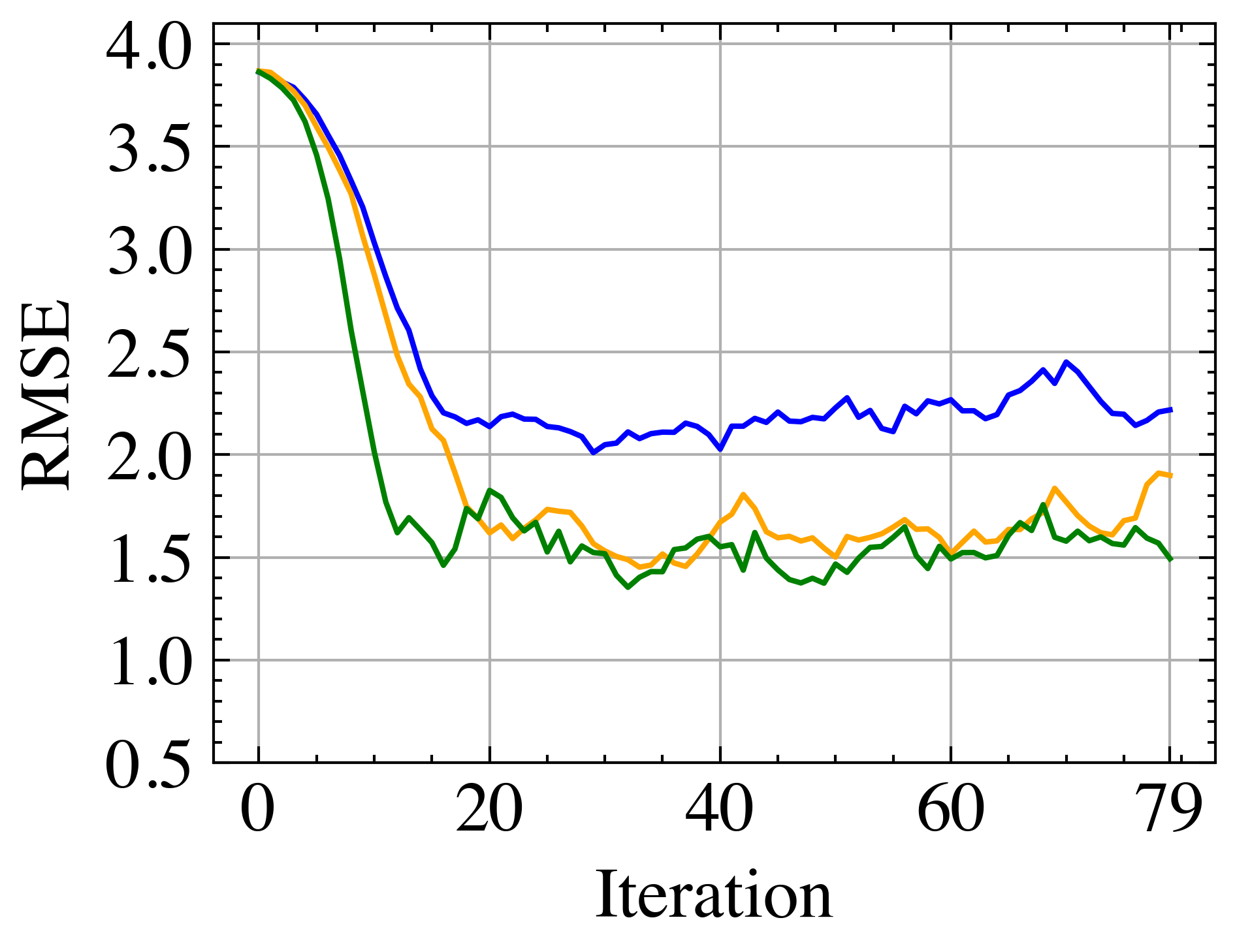}
            \caption{Item = 70, User = 15}
            \label{fig:s70}
        \end{subfigure}
        \hspace{-0.2cm}
        \begin{subfigure}[c]{0.33\textwidth}
            \centering
            \includegraphics[width=\linewidth]{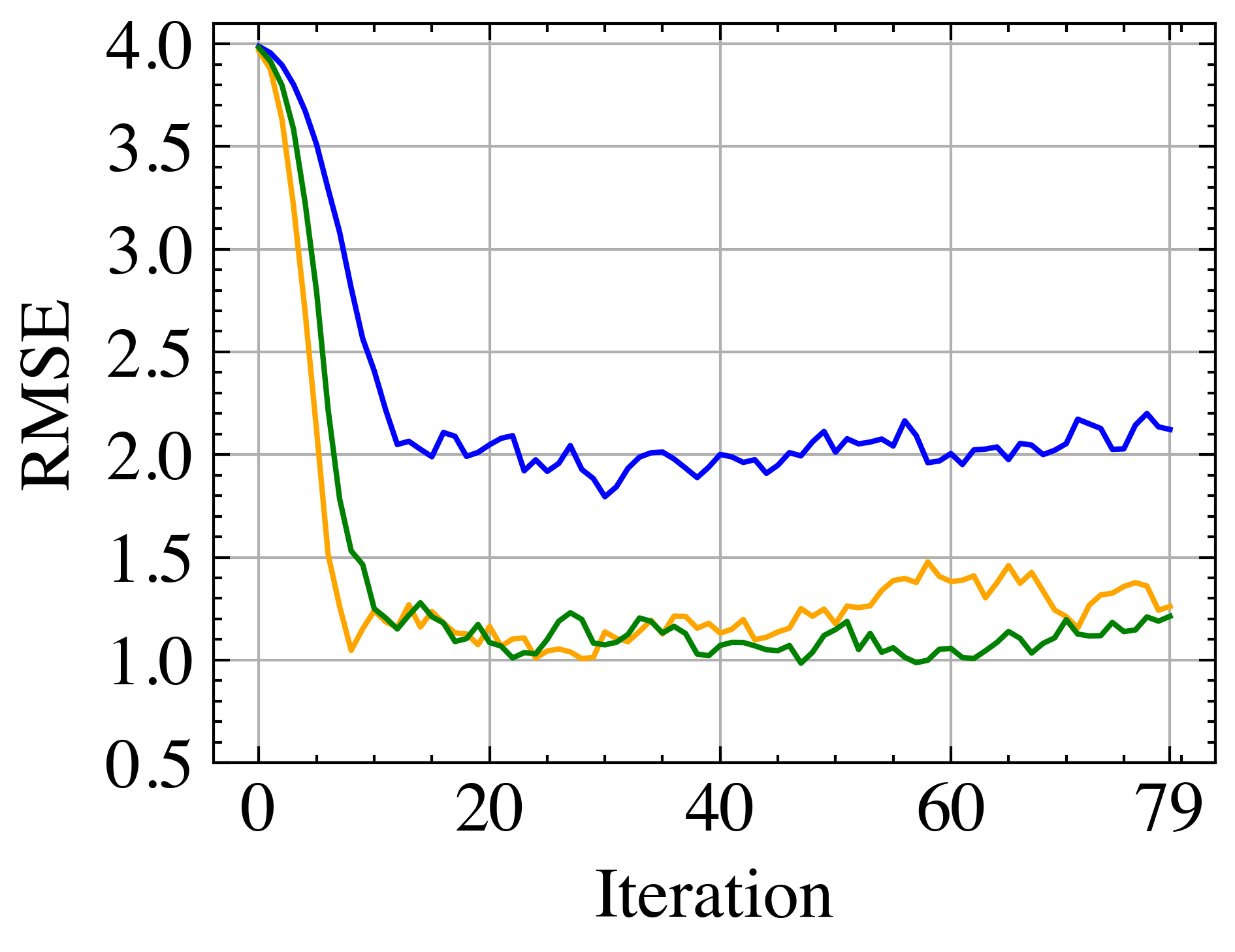}
            \caption{Item = 90, User = 20}
            \label{fig:s90}
        \end{subfigure}
    \end{minipage}
    \hspace{-0.3cm}
    \begin{minipage}{0.10\linewidth}
        \centering
        \begin{tikzpicture}
            \draw[blue, line width=1pt] (0,5) -- (0.5,5);
            \node[anchor=west] at (0.6,5) {10$\%$};

            \draw[myyellow, line width=1pt] (0,4.5) -- (0.5,4.5);
            \node[anchor=west] at (0.6,4.5) {20$\%$};

            \draw[mygreen, line width=1pt] (0,4) -- (0.5,4);
            \node[anchor=west] at (0.6,4) {30$\%$};
        \end{tikzpicture}
    \end{minipage}
    \caption{RMSE results of varied data sharing fraction and training dataset sizes in LIBERATE.}
    \label{fig:3ginshare}
\end{figure*}
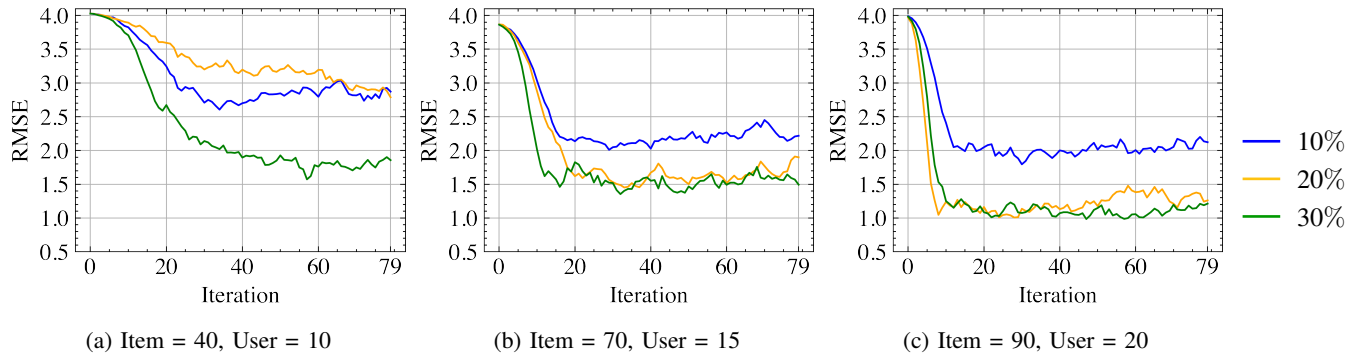

Fig. \ref{fig:3ginshare} elucidates the influence of the shared data fraction on the model's RMSE across three distinctive graphs. The blue, yellow, and green lines in these graphs respectively represent the extensions of the local dataset by 10$\%$, 20$\%$, and 30$\%$. Fig. 6(a), (b), and (c) display results under varying dataset complexities, including 40 items with 10 users, 70 items with 15 users, and 90 items with 20 users respectively. As observed in these figures, a larger fraction of shared data, even in the presence of differential privacy, can enhance the model's convergence speed.

\begin{table}[h]
\centering
\caption{\label{Tab: rmsefc} The influence of data sharing fraction on final RMSE.}
\begin{tabular}{lccc}
\toprule
\multicolumn{1}{c}{\multirow{2}{*}{Fraction}} & \multicolumn{3}{c}{Dataset Size}                             \\ \cline{2-4} 
\multicolumn{1}{c}{}                               & \multicolumn{1}{c}{$U$ = 10, $I$ = 40} & \multicolumn{1}{c}{$U$ = 15, $I$ = 70} & \multicolumn{1}{c}{$U$ = 20, $I$ = 90} \\ \midrule
10\% Share                                         & 2.879 $\pm$ 0.096             & 2.210 $\pm$ 0.091            & 2.109 $\pm$ 0.092             \\
20\% Share                                         & 2.795 $\pm$ 0.093            & 1.899 $\pm$ 0.079             & 1.261 $\pm$ 0.066            \\
30\% Share                                         & 1.803 $\pm$ 0.074              & 1.499 $\pm$ 0.07             & 1.137 $\pm$ 0.053           \\ \bottomrule
\end{tabular}

\end{table}

Table \ref{Tab: rmsefc} presents an insightful analysis of how the extent of data sharing and the volume of training data interact to affect the final RMSE in LIBERATE. Within Table \ref{Tab: rmsefc}, $U$ represents the number of users, and $I$ represents the number of items. Across all evaluated dataset sizes, there appears to be an inverse correlation between the fraction of shared data and the RMSE. This trend suggests that enhancing the data-sharing fraction contributes to improved model performance, as represented by a lower RMSE. Consider the case when the dataset consists of 10 users and 40 items. Here, an escalation in the data sharing fraction from 10 $\%$ to 30 $\%$ culminates in a decrease in the RMSE from approximately 2.879 to 1.803. This observation implies that a higher sharing fraction, and consequently more data for model training, leads to better model performance. Furthermore, the influence of the dataset size on RMSE is clearly evident. As we progress from smaller to larger datasets, the RMSE displays a general decline, implying improved model performance with increasing data volume. This trend persists even when the data-sharing fraction remains constant. As an example, with a fixed data sharing a fraction of 20 $\%$, the RMSE drops from roughly 2.795 (for a dataset of 10 users and 40 items) to about 1.261 (for a dataset with an enhanced scale of 20 users and 90 items).

\begin{figure}[h]
\centering
\captionsetup{justification=centering}
\includegraphics[scale = 0.71]{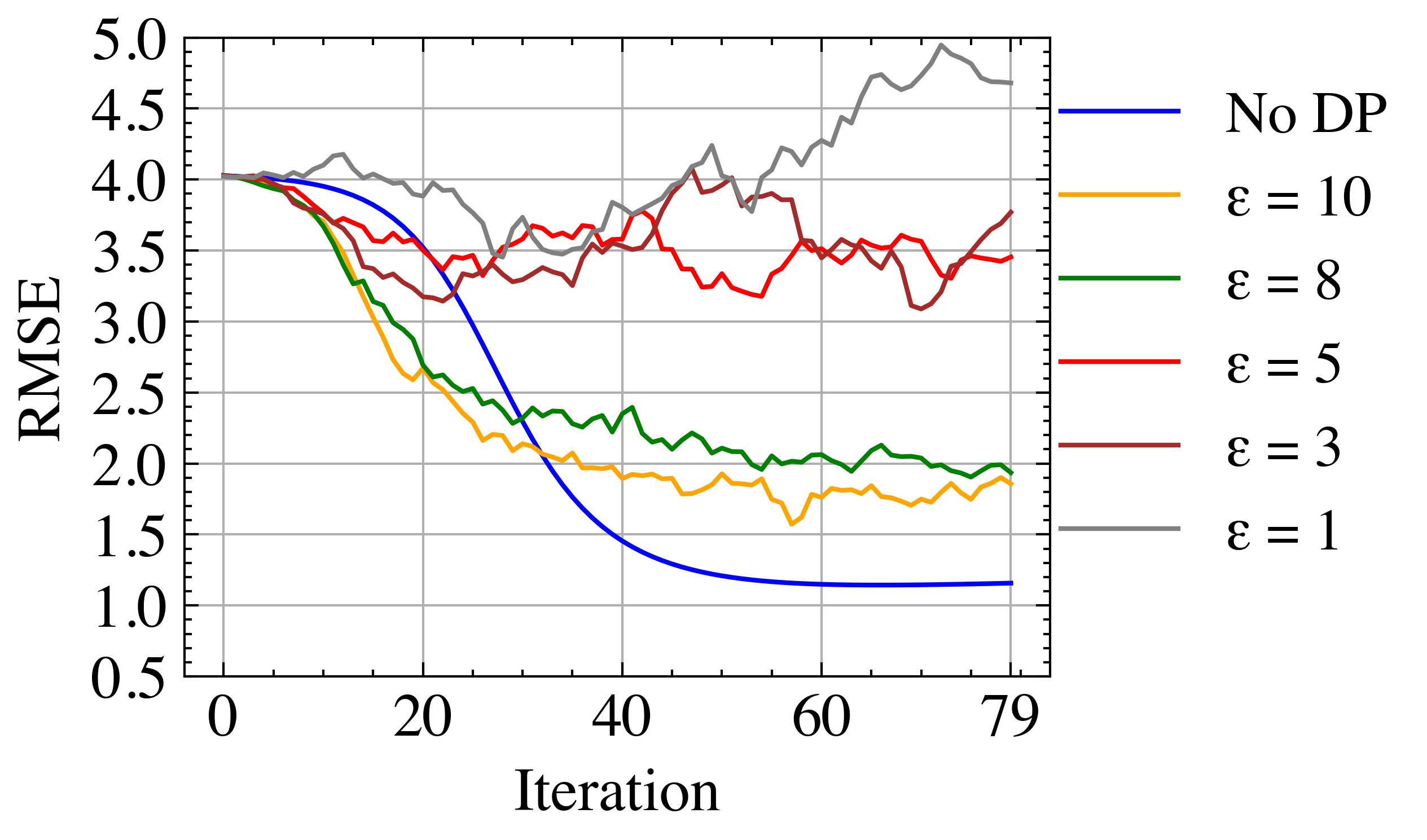}
\caption{RMSE results of LIBERATE with different privacy budget $\epsilon$.}
\label{fig:rmsedp}
\end{figure}

Fig. \ref{fig:rmsedp} elucidates the effects of deploying differential privacy, quantified by the privacy budget parameter $\epsilon$, on the performance of the LIBERATE model. The results derived from this figure indicate an inverse relationship between $\epsilon$ and the model's performance. In our experimental study, we scrutinized the performance of LIBERATE under varying values of $\epsilon$, specifically 1, 3, 5, 8, and 10. A pattern emerges from Fig \ref{fig:rmsedp}: as $\epsilon$ decreases, representing an increase in the level of privacy and consequently, a higher noise injection, the RMSE correspondingly increases. This trend suggests that smaller values of $\epsilon$ are accompanied by a larger error, thereby diminishing model performance during training. The most optimal performance of LIBERATE is observed when differential privacy is not implemented. When the privacy budget parameter $\epsilon$ falls between 5 and 10, the RMSE of LIBERATE witnesses an increase of approximately 0.6 to 0.8 compared to LIBERATE without the integration of differential privacy. However, it is crucial to consider that the observed non-convergence in model performance under differential privacy could be primarily attributed to the small size of the training dataset. This suggests that with a larger dataset, the effect of differential privacy on model performance could potentially be less detrimental.

\begin{figure*}[h]
    \centering
    \begin{minipage}{0.85\linewidth}
        \begin{subfigure}[c]{0.33\textwidth}
            \centering
            \includegraphics[width=\linewidth]{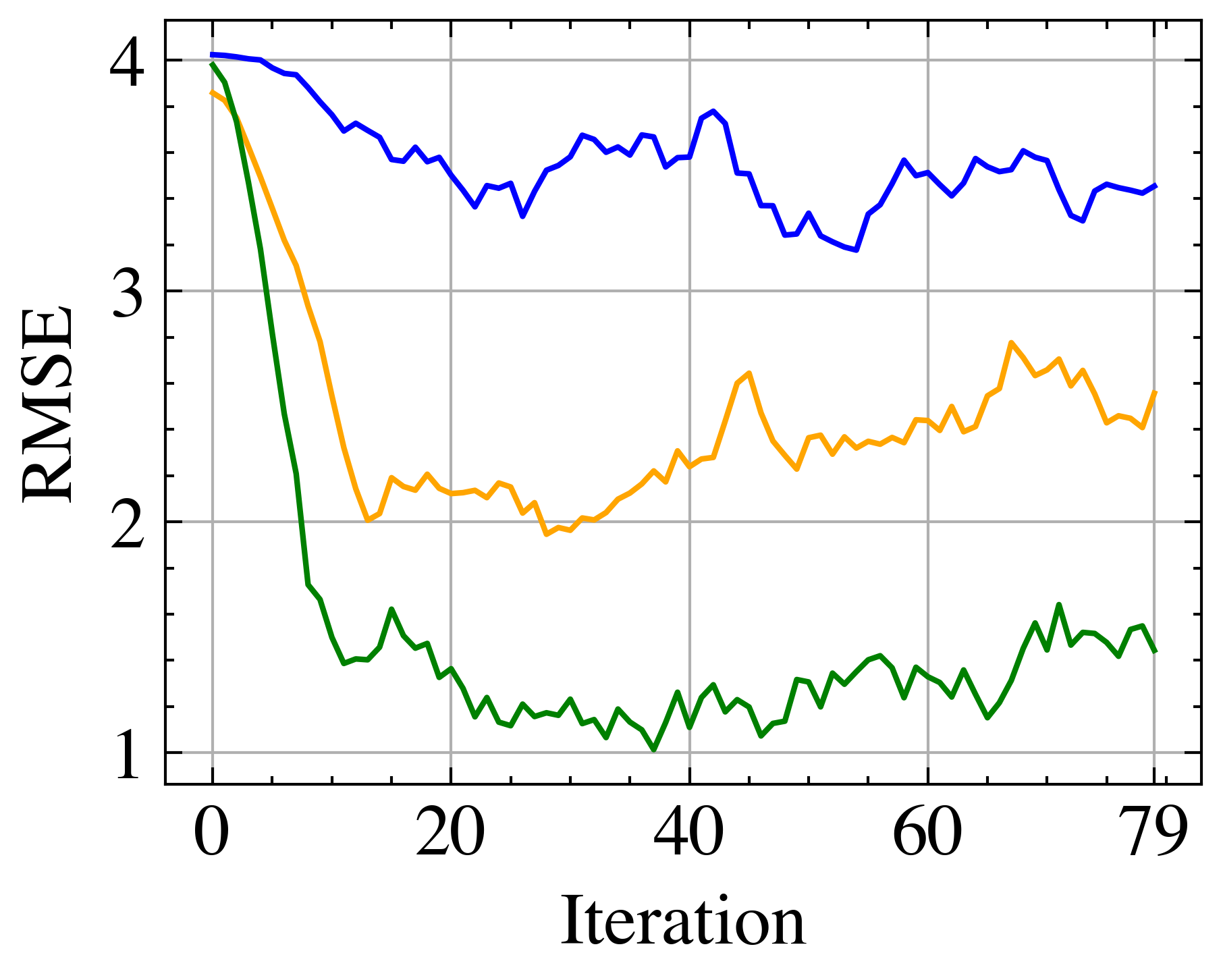}
            \caption{$\epsilon = 5$}
            \label{fig:dp5}
        \end{subfigure}
        \hspace{-0.2cm}
        \begin{subfigure}[c]{0.33\textwidth}
            \centering
            \includegraphics[width=\linewidth]{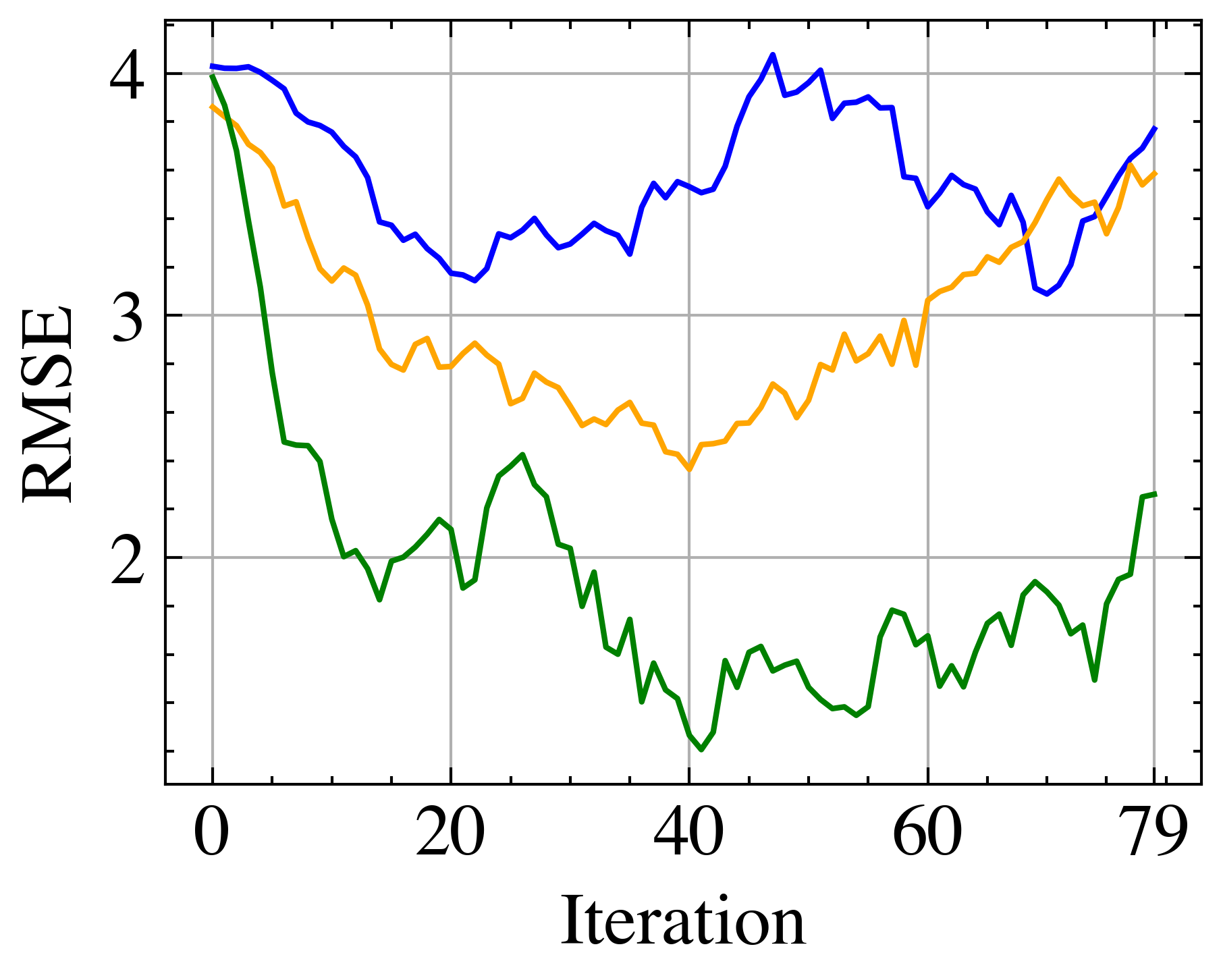}
            \caption{$\epsilon = 3$}
            \label{fig:dp3}
        \end{subfigure}
        \hspace{-0.2cm}
        \begin{subfigure}[c]{0.33\textwidth}
            \centering
            \includegraphics[width=\linewidth]{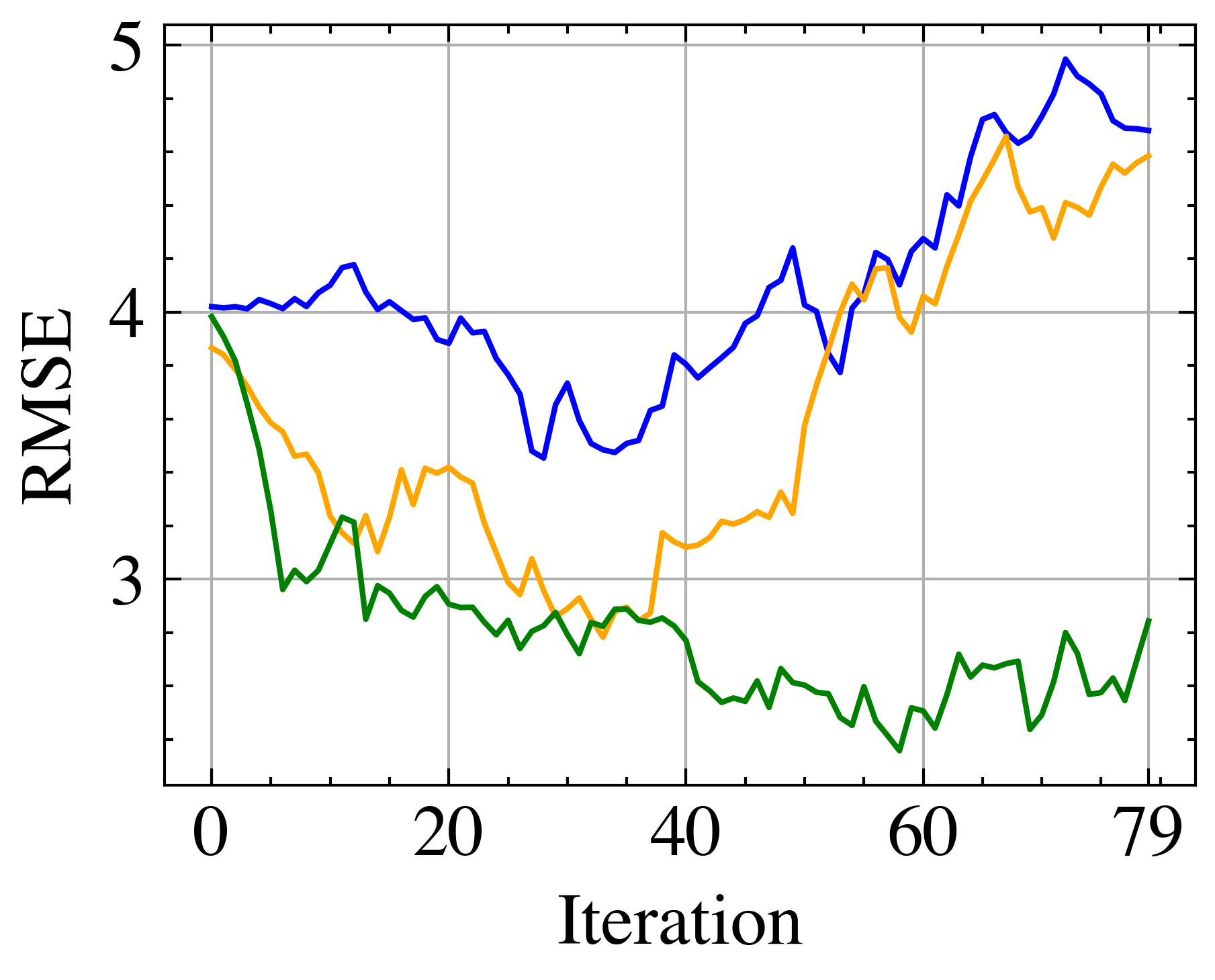}
            \caption{$\epsilon = 1$}
            \label{fig:dp1}
        \end{subfigure}
    \end{minipage}
    \hspace{-0.3cm}
    \begin{minipage}{0.15\linewidth}
        \centering
        \begin{tikzpicture}
            \draw[blue, line width=1pt] (0,5) -- (0.5,5);
            \node[anchor=west] at (0.6,5) {U=10, I=40};

            \draw[myyellow, line width=1pt] (0,4.5) -- (0.5,4.5);
            \node[anchor=west] at (0.6,4.5) {U=15, I=70};

            \draw[mygreen, line width=1pt] (0,4) -- (0.5,4);
            \node[anchor=west] at (0.6,4) {U=20, I=90};
        \end{tikzpicture}
    \end{minipage}
    \caption{Influence of differential privacy on LIBERATE's performance across varied training dataset sizes.}
    \label{fig:three graphs}
\end{figure*}

Fig. \ref{fig:three graphs} presents the performance of LIBERATE under varying dataset sizes and differential privacy levels, denoted by the privacy budget parameter, $\epsilon$. Each of the three graphs portrays the model's performance under a distinct $\epsilon$ value: 5 in Fig. 8(a), 3 in Fig. 8(b), and 1 in Fig. 8(c). Each chart comprises three lines, each representing a different dataset size. The blue line corresponds to a dataset size of 10 users and 40 items; the yellow line corresponds to a dataset of 15 users and 70 items; and the green line represents a dataset size of 20 users and 90 items. As previously noted, when $\epsilon$ falls below 5 and the training dataset is small, the model tends not to converge due to the high level of noise injected by differential privacy. To assess the model's performance when $\epsilon$ is below 5, we consequently increased the size of the training dataset, examining sizes of 15 users with 70 items and 20 users with 90 items. Fig. 8(c) indicates that the disruptive impact of differential privacy on model performance diminishes as the dataset size increases. Notably, even when $\epsilon$ is set to 1, implying a high level of privacy and noise, LIBERATE still converges with a training dataset of 20 users and 90 items. This implies that the model retains its capacity to provide suitable recommendation results while preserving privacy, when the dataset size is sufficiently large.

\subsection{Time Complexity Analysis} 

\begin{table*}[b]
\centering
\caption{\label{Tab: time}Time consumption of baselines (seconds).}
\begin{tabular}{lccc}
\toprule
\multicolumn{1}{c}{\multirow{2}{*}{Model}} & \multicolumn{3}{c}{Time Consumption}                                     \\ \cline{2-4} 
\multicolumn{1}{c}{}                       & Communication Time & Blockchain Time  & Overall Time \\ 
\midrule
SecureFedMF($U$ = 10, $I$ = 40, Without HE)          & 16.395           & -                            & 16.395    \\
SecureFedMF($U$ = 10, $I$ = 40, With HE)                & 594.674            & -                            & 594.674      \\
LIBERATE($U$ = 10, $I$ = 40, Without DP)        & 23.404             & 33.518                        & 56.922       \\
LIBERATE($U$ = 10, $I$ = 40, $\epsilon$ = 10)           & 23.537             & 33.518                        & 57.055       \\
LIBERATE($U$ = 15, $I$ = 70, Without DP)        & 37.800             & 117.224                       & 155.024      \\
LIBERATE($U$ = 15, $I$ = 70, $\epsilon$ = 10)           & 37.923             & 117.224                       & 155.147      \\
LIBERATE($U$ = 20, $I$ = 90, Without DP)        & 50.238             & 359.635                      & 409.873      \\
LIBERATE($U$ = 20, $I$ = 90, $\epsilon$ = 10)           & 50.937             & 359.635                       & 410.527      \\ 
\bottomrule
\end{tabular}

\end{table*}

\begin{figure}[h]
\centering
\captionsetup{justification=centering}
\includegraphics[scale = 0.475]{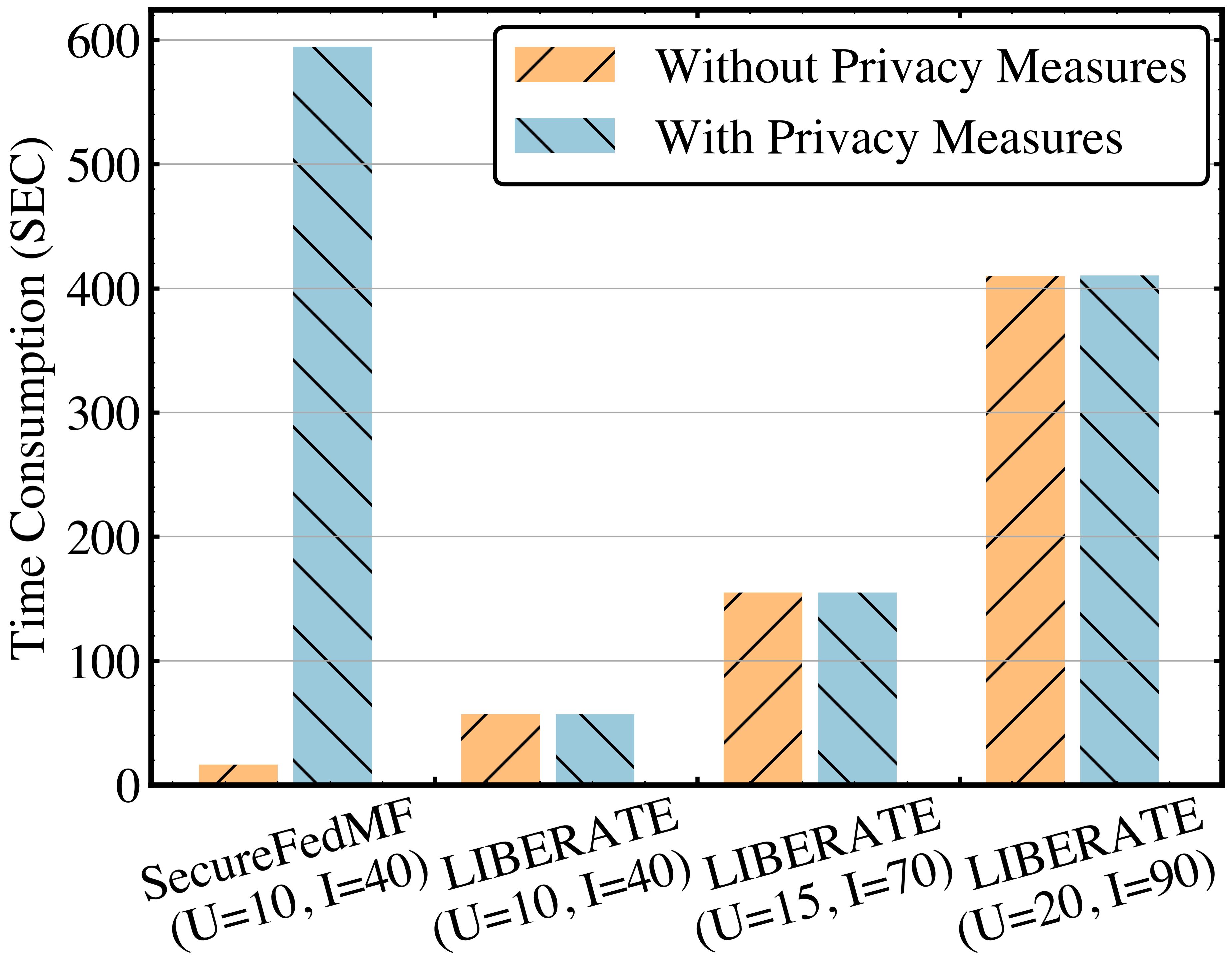}
\caption{Time efficiency comparison: time consumption between LIBERATE and Secure FedMF in terms of Blockchain processing and user-server communication.}
\label{fig:timebar}
\end{figure}

In the analysis of computational time efficiency, we investigated the operational timing of LIBERATE, focusing on transaction recording time within the blockchain framework and the communication time between users and the server. It should be noted that the integration of the blockchain does not impede the baseline model's training time. Therefore, our analysis focuses solely on the communication and blockchain-related timing. The results of this investigation are compiled in Table \ref{Tab: time} and Fig. \ref{fig:timebar}. Within Table \ref{Tab: time} and Fig. \ref{fig:timebar}, $U$ represents the number of users and $I$ represents the number of items. 

Fig. \ref{fig:timebar} illustrates the time consumption of both LIBERATE and Secure FedMF with and without privacy-preserving measures. Secure FedMF is executed with 10 users and 40 items. Conversely, LIBERATE is tested under various user counts (10, 15, and 20) and item counts (40, 70, and 90). In scenarios without privacy-preserving techniques, the time consumption for SecureFedMF is significantly less than LIBERATE. This elevated time consumption for LIBERATE can be attributed to its utilization of blockchain for traceable data sharing and model updates. It's noteworthy that as the number of users scales up in the LIBERATE model, there is a concomitant rise in time consumption. However, this increase remains within an acceptable range. When privacy-preserving mechanisms are introduced, the disparities between the models become more pronounced. SecureFedMF, which adopts homomorphic encryption, witnessed a drastic surge in its time consumption. This can be directly linked to the computationally intensive nature of homomorphic encryption, which requires significant time and resources. In contrast, LIBERATE employs differential privacy, which would not significantly exacerbate time consumption.

Table \ref{Tab: time} lists the detailed time consumption of LIBERATE and Secure FedMF. Our selection of differential privacy as a safeguard against information leakage from gradient uploads has resulted in a reduction of communication time between the server and the user in comparison to Secure FedMF. As evidenced in Table \ref{Tab: time}, the communication time for LIBERATE escalates incrementally as the dataset size enlarges. Despite this increase, the addition of differential privacy does not exert a significant influence on the final communication time, even when large datasets are utilized. This implies that LIBERATE can deliver efficient operational performance while preserving privacy, even in the context of large datasets.

The most resource-intensive component of our model is the blockchain, where data-sharing transactions are recorded on each block. This is primarily attributable to the computational complexity of the hash function. As the number of users and items swells, so does the time expenditure on the blockchain for transaction recording. For instance, the blockchain operation time reaches nearly 360 seconds for a training dataset comprising 90 items and 20 users. In contrast, the blockchain time is a mere 33 seconds when the training dataset consists of 40 items and 10 users.

\begin{table}[h]
\centering
\caption{\label{Tab: hashdf}Hash difficulty level with time consumption.}
\resizebox{\linewidth}{0.0565\textheight}{
\begin{tabular}{cccc}
\toprule
\multirow{2}{*}{Difficulty   Level} & \multicolumn{3}{c}{Dataset Size}                 \\ \cline{2-4} 
                                    & $U$ = 10, $I$ = 40 & $U$ = 15, $I$ = 70 & $U$ = 20, $I$ = 90 \\ \midrule
1                                   & 0.0012         & 0.0051         & 0.0135         \\
2                                   & 0.0177         & 0.0441         & 0.6518         \\
3                                   & 0.1105         & 0.8228         & 1.1813         \\
4                                   & 1.6149         & 4.9770         & 17.0530        \\
5                                   & 33.5180        & 117.2242       & 359.6348       \\ \bottomrule
\end{tabular}
}
\end{table}

\begin{figure}[h]
\centering
\captionsetup{justification=centering}
\includegraphics[width=\linewidth]{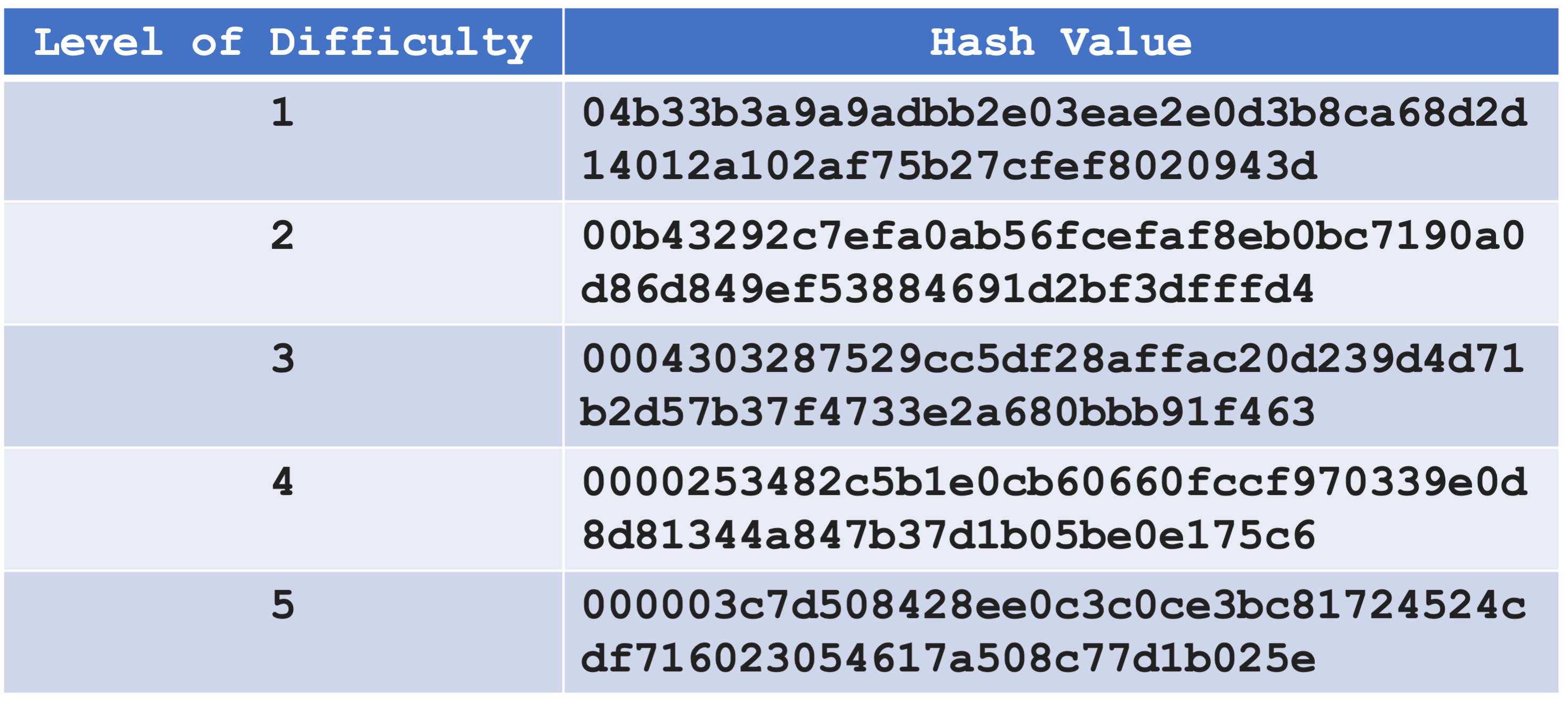}
\caption{Example of hash value corresponding to different levels of computational difficulty.}
\label{fig:hash}
\end{figure}

Table \ref{Tab: hashdf} presents the correlation between the difficulty level of hashing and time consumption in the context of the LIBERATE model. Within Table \ref{Tab: hashdf}, $U$ represents the number of users, and $I$ represents the number of items. In LIBERATE, the difficulty level corresponds to the number of leading zeros in the hash value. In LIBERATE, the degree of difficulty of the hash function during the creation of a new block in the proposed blockchain plays a vital role in determining the time efficiency. The difficulty level is determined by the number of leading zeros in the hash value. The more zeros, the greater the complexity in mining, thereby enhancing the security for information storage. For comparison, Bitcoin, one of the most widely used digital currencies, operates with a hash function incorporating 23 leading zeros. Evidently, as the difficulty level escalates, the time consumption also increases, irrespective of the dataset size. For a user-item configuration of 10 and 40, time consumption ascends from a mere 0.0012 seconds at difficulty level 1 to a significant 33.5180 seconds at level 5. A similar pattern is observed for larger dataset sizes with more users and items, signifying a direct relationship between the difficulty level and time consumption. A visual demonstration of the hash value under varying levels of difficulty during our experiment is exhibited in Fig. \ref{fig:hash}.

Another noteworthy observation from Table \ref{Tab: hashdf} is the influence of the dataset size on the time consumption when operating under the same difficulty level. Evidently, as the dataset size escalates, the time spent on operations also increases, illustrating a positive correlation between these two variables. For example, at difficulty level 1, the time consumption for the user-item configuration of 10 and 40 is 0.0012 seconds. However, when the dataset size increases to 15 users with 70 items and 20 users with 90 items, the time consumption rises to 0.0051 seconds and 0.0135 seconds, respectively. This trend is consistent across all difficulty levels, as can be observed from the table. At the highest difficulty level of 5, the time consumption for the user-item configurations of 10 and 40, 15 and 70, and 20 and 90 are 33.5180 seconds, 117.2242 seconds, and 359.6348 seconds, respectively.

\subsection{Privacy Analysis}

In the design of LIBERATE, we incorporate a blockchain-based traceability mechanism for recording the detailed status of data sharing and model updates. This utilization of blockchain enables users the ability to seamlessly track the distribution and reception of their data, as well as comprehend the processes of model updating. When a user encounters privacy concerns from potential malicious activities, the user is equipped with the means to investigate the diffusion path of their data, updates of both local and server-side models. Furthermore, they can identify which users have contributed data for training purposes to their devices. This section presents a comprehensive case study on illustrating the traceability of data sharing and model updates within LIBERATE.

Consider the scenario of an active user, Alice, who frequently engages with various smart devices within IoT. Alice relies on the federated recommender system for her daily needs. Alice found herself entangled in threads regarding data privacy. Guided by the privacy-preserving strategy of LIBERATE, Alice embarked on a mission to unravel the journey of her shared data across the vast network of interconnected IoT devices. In the secure and transparent archives of the blockchain, Alice sought clarity, navigating through the immutable records that narrate the tale of her data’s path. As she traversed the blockchain’s records, Alice discovered the realms where her data had been utilized and the entities that had been guardians of her shared essence. Armed with the power of traceability, she unveiled the cycles of model updates influenced by her data, gaining insights into the stages of its utilization and the consequential impact on recommendation processes.

Alice have also noticed about malicious entities trying to manipulate model updates to steal private data. With LIBERATE's blockchain-based traceability mechanism, every model updates were transparently recorded. By reviewing the blockchain history, Alice could see all changes to the model. This allowed her to spot any unusual patterns or potential unauthorized changes, ensuring her data remained privacy. This traceability gave Alice confidence that her private information was protected from malicious intents.


Indeed, Alice's exploration of the blockchain doesn't stop at understanding her recommendations. As she delves deeper into the blockchain records, she observes a suspicious pattern: a large number of users have recently provided unusually high ratings for a product that typically does not receive such positive feedback. Intrigued by this abnormal trend, Alice decides to investigate. Leveraging the traceability and transparency offered by LIBERATE's blockchain implementation, Alice tracks the origin of these suspicious ratings. She observes that these ratings first appeared from a small group of users and then spread quickly across the system, influencing the item profile matrix $V$ and thereby the recommendations received by other users. Upon further investigation, Alice discovers that these users are in fact 'bot' accounts. The patterns in their behavior, such as the frequency and timing of their ratings, coupled with the uniformity of their high ratings, signal a coordinated effort to manipulate the recommender system artificially. This manipulation could potentially harm the security and performance of the recommender system, affecting the quality of recommendations received by other users. Equipped with her findings, Alice reports the malicious behavior of these users to the system administrators. Because the blockchain records are immutable and traceable, Alice's claims are easily verifiable. Consequently, the system administrators promptly take action, removing the suspicious ratings from the system and banning the identified bot accounts to prevent further poisoning to the recommendation model.

This additional incident showcases the crucial role of LIBERATE's blockchain implementation in detecting and mitigating malicious activities. By leveraging traceability, LIBERATE allows users to observe and report suspicious behaviors, protecting the privacy of data sharing and model updates.

\subsection{Discussion}
This section delineates current limitations inherent in  LIBERATE, and outlines potential directions for future research aimed at enhancing its practicality and effectiveness in real-world applications.

\textit{Enhancing Blockchain Efficiency.} As evidenced in the second part of our evaluation, the time required for the hash function computation within the blockchain increases proportionally with the growth in the number of users and items. Given the complexity of the hash function, the cumulative time expenditure on the blockchain can result in significant delays, which could impede the timely delivery of recommendations in scenarios where numerous interconnected devices are engaged in the recommendation system. Hence, future work focusing on the improvement of blockchain computational efficiency would be highly advantageous for our recommendation system.

\textit{Balancing Security, Accuracy, and Efficiency in Highly Sensitive Contexts.} This study is predicated on the assumption that the training data used for recommendations is permitted to be shared amongst interconnected devices, thus facilitating the use of blockchain for data sharing records. However, in certain industry contexts, such as healthcare or banking, user data is classified as highly sensitive, and therefore, data sharing becomes untenable. The usage of differential privacy in these contexts may impinge on the predictive capability of the model. While cryptographic methods can be applied as a potential solution, they may incur significant computational costs and time, particularly considering the limited computational resources typical of IoT-connected devices. Future research will need to explore strategies for reconciling efficiency, privacy, and computational complexity, with the objective of providing an optimal solution for recommendation services.

\section{Conclusion}\label{sec6}


In this study, we introduced LIBERATE, a matrix factorization-based federated recommender system that employs a blockchain-based traceability mechanism for marking the pace of data sharing and model updates, thereby providing a privacy-preserving solution. Within LIBERATE, details of data-sharing and model updates are recorded into a blockchain. The implementation of this blockchain-based traceability mechanism ensures the protection of data privacy during data sharing and model updates. This method enabling users to trace the origin and destination of their data. Moreover, users can easily track the model's updates history, pinpointing the exact nodes where model parameters were potentially malicious. This, in turn, bolsters the privacy-preserving of the federated recommender system, giving users increased confidence in employing this system. Local differential privacy is employed to ensure high-level privacy protection of the model updates between the server and clients. Experiments on a real-world dataset illustrate that LIBERATE outperforms comparable systems in model performance. Furthermore, LIBERATE offers substantial benefits for improving the trustworthiness of the recommendations, as demonstrated in our case study. 



LIBERATE exhibits a substantial step forward in the design of privacy-preserving for data sharing and model updates within federated recommender systems. In subsequent research, we will focus on enhancing computational efficiency and developing solutions to address the challenges in high-sensitivity data contexts, thereby further extending the applicability of this promising framework.

\section*{ACKNOWLEDGEMENTS}
The authors would like to express their profound gratitude to Zhiwen Han and Huafei Huang, from the Dalian University of Technology for their support to this research.

%


\bibliographystyle{IEEEtran}
\bibliography{bare_jrnl_new_sample4}


\vspace{-15pt}

\begin{IEEEbiography}[{\includegraphics[width=1in,height=1.25in,clip,keepaspectratio]{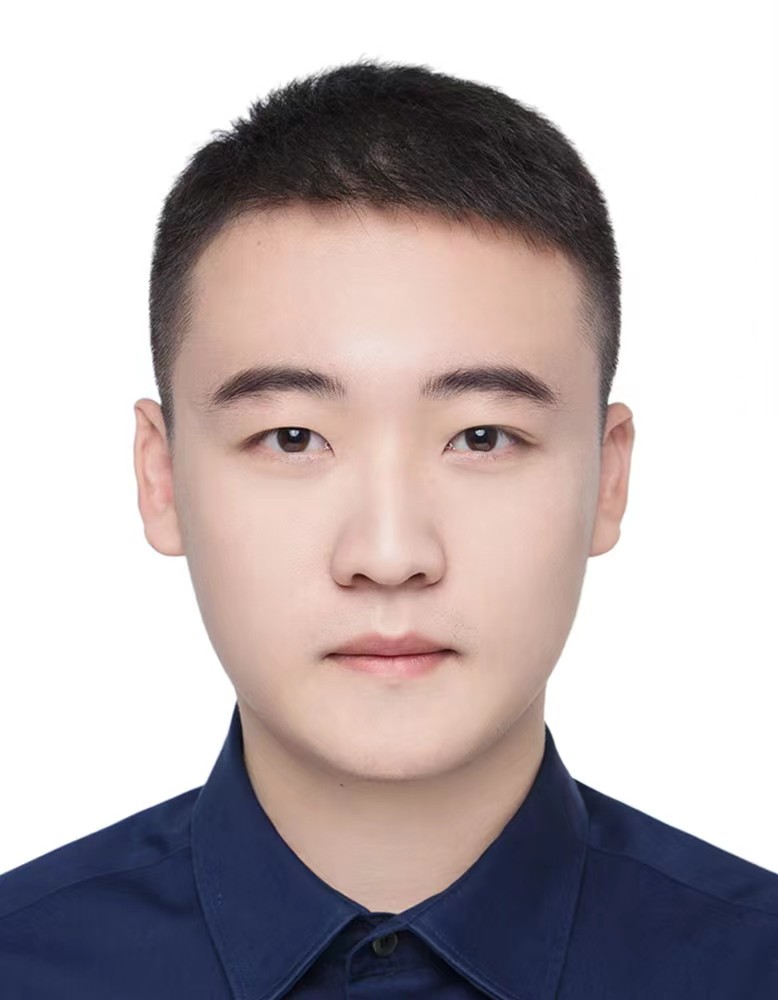}}]{Zhen Cai}
(Student Member, IEEE) is a Ph.D. candidate specializing in Information Technology at the Institute of Innovation, Science, and Sustainability, Federation University, Australia. He completed his Masters degree in Business Analytics at Deakin University, Australia in 2021. Mr. Cai's research primarily focuses on decentralized, privacy-preserving recommender systems. His academic interests includes, but not limited to, business analytics, recommender systems, and the ethical implications of artificial intelligence. 

\end{IEEEbiography}

\vspace{10pt}

\begin{IEEEbiography}[{\includegraphics[width=1in,height=1.25in,clip,keepaspectratio]{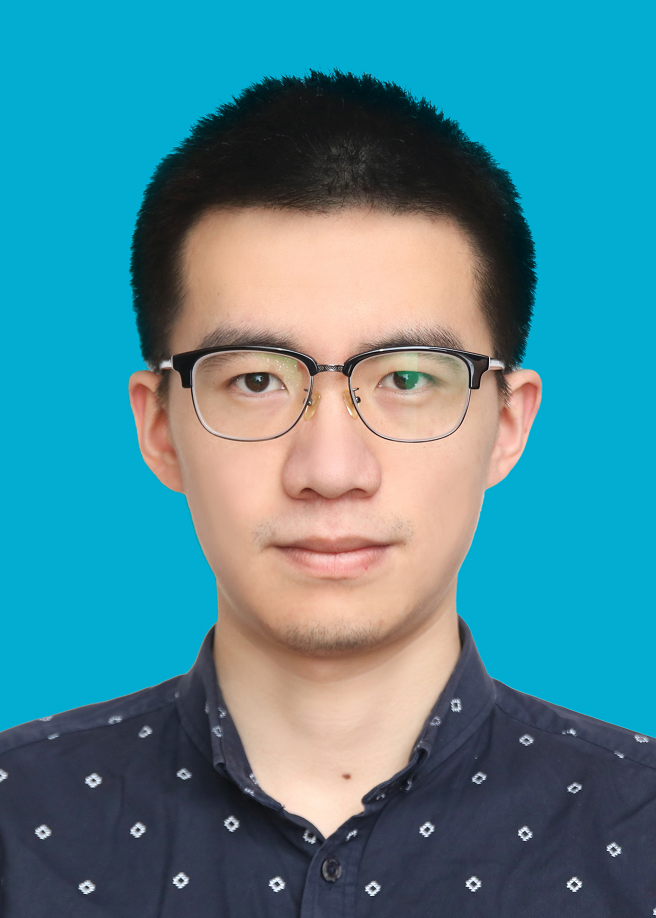}}]{Tang Tao} 

(Student Member, IEEE) received the Bachelor's Degree from the Chengdu College, University of Electronic Science and Technology of China, Chengdu, China in 2019. He is currently a PhD student at the Institute of Innovation, Science, and Sustainability, Federation University Australia. His research interests include graph learning, big data analytics, recommender systems, and computational intelligence.

\end{IEEEbiography}

\vspace{11pt}

\begin{IEEEbiography}[{\includegraphics[width=1in,height=1.25in,clip,keepaspectratio]{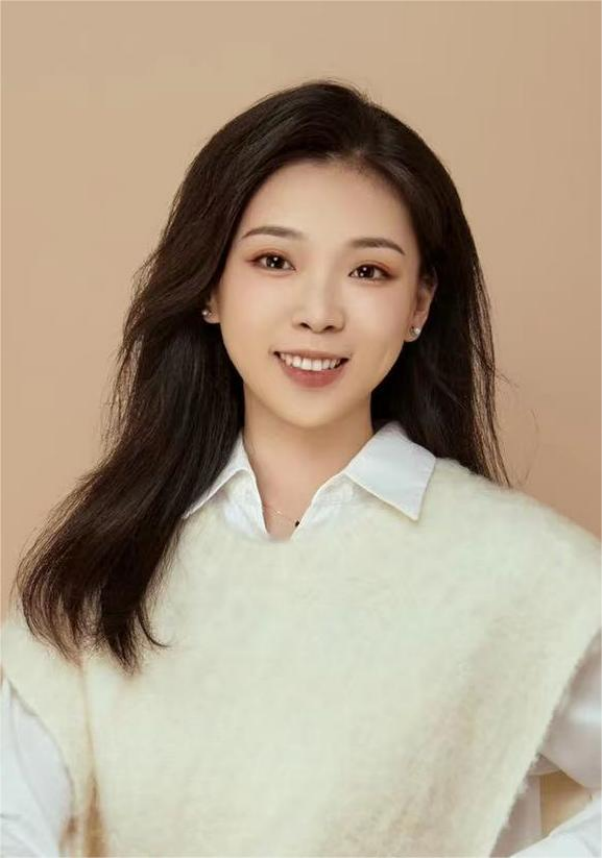}}]{Shuo Yu} 

 (M’20) received B.Sc. and M.Sc. degrees from Shenyang University of Technology, Shenyang, China. She received her Ph.D. degree from Dalian University of Technology, Dalian, China. She received her Ph.D. degree from Dalian University of Technology. She is currently an Associate Professor in School of Computer Science and Technology, Dalian University of Technology. She has published over 50 papers in ACM/IEEE conferences, journals, and magazines, and received several academic awards, including IEEE DataCom 2017 Best Paper Award, IEEE CSDE 2020 Best Paper Award, and ACM/IEEE JCDL 2020 The Vannevar Bush Best Paper Honorable Mention. She has served as the Track Chair and PC member of three international conferences. Her research interests include network science, data science, computational social science, and knowledge science. She is a Member of IEEE.

\end{IEEEbiography}

\vspace{11pt}

\begin{IEEEbiography}[{\includegraphics[width=1in,height=1.25in,clip,keepaspectratio]{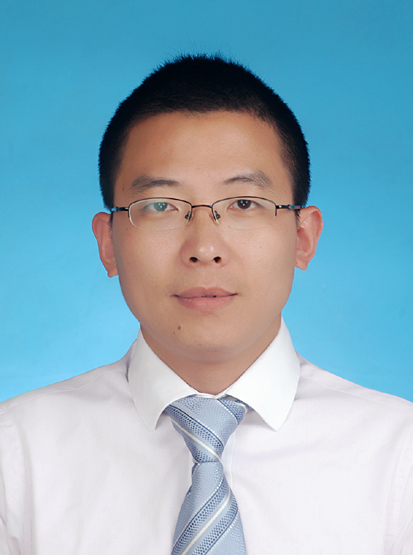}}]{Yunpeng Xiao} 

received the Ph.D. degree in computer science from Beijing University of Posts and Telecommunications, Beijing, China, in 2013. He is a professor and vice president of the Institute of Electronic Information and Network Engineering, Chongqing University of Posts and Telecommunications, Chongqing, China. He was a visiting scholar of Georgia Institute of Technology from 2018 to 2019. His research interests include social networks, e-commerce and intelligent systems.

\end{IEEEbiography}

\begin{IEEEbiography}[{\includegraphics[width=1in,height=1.25in,clip,keepaspectratio]{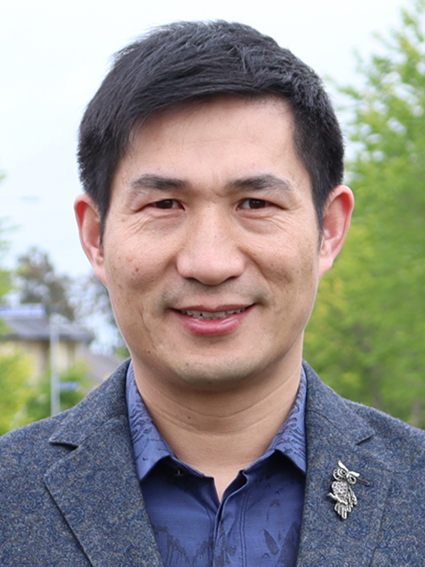}}]{Feng Xia} 

(M’07-SM’12) received the BSc and PhD degrees from Zhejiang University, Hangzhou, China. He is a Professor in School of Computing Technologies, RMIT University, Australia. Dr. Xia has published 2 books and over 300 scientific papers in international journals and conferences (such as IEEE TAI, TKDE, TNNLS, TC, TMC, TPDS, TBD, TCSS, TNSE, TETCI, TETC, THMS, TVT, TITS, TASE, ACM TKDD, TIST, TWEB, TOMM, WWW, AAAI, SIGIR, WSDM, CIKM, JCDL, EMNLP, and INFOCOM). His research interests include data science, artificial intelligence, graph learning, and systems engineering. He is a Senior Member of IEEE and ACM, and an ACM Distinguished Speaker.

\end{IEEEbiography}

\vspace{-40pt}



\end{document}